\documentclass[11pt]{article}
\usepackage[preprint]{acl}

\usepackage{caption}
\usepackage{needspace}
\usepackage{times}
\usepackage{latexsym}
\usepackage{enumitem}
\usepackage{amsmath}
\usepackage{amssymb}
\usepackage{tabularx}
\usepackage[table]{xcolor}
\usepackage[ruled]{algorithm2e}
\definecolor{lightblue}{RGB}{220,235,250}
\usepackage{pifont}
\usepackage{xcolor,colortbl}
\usepackage{caption}

\usepackage[T1]{fontenc}

\usepackage[utf8]{inputenc}
\usepackage{multirow}
\usepackage{adjustbox}
\usepackage{hyperref}
\usepackage{float}
\usepackage{siunitx}
\usepackage{cuted}
\usepackage{caption}
\usepackage{booktabs}
\usepackage{placeins}
\usepackage{xspace}

\usepackage{pifont}
\usepackage[table]{xcolor}

\usepackage{microtype}

\usepackage{inconsolata}

\usepackage{graphicx}
\usepackage{booktabs}
\usepackage{multirow}
\usepackage{multicol}
\usepackage{subcaption}
\usepackage{amsmath}
\usepackage[most,skins,theorems]{tcolorbox}
\usepackage{xcolor}

\definecolor{lightpink}{RGB}{238,206,205}
\definecolor{grayhighlight}{RGB}{240,240,240}
\definecolor{lightblue}{RGB}{210, 222, 238}

\tcbset{
  aibox/.style={
    width=\linewidth,
    top=8pt,
    bottom=4pt,
    colback=lightpink,
    colframe=black,
    colbacktitle=black,
    enhanced,
    center,
    attach boxed title to top left={yshift=-0.1in,xshift=0.15in},
    boxed title style={boxrule=0pt,colframe=white,},
  }
}

\title{LaRA: Layer-wise Representation Analysis for \\ Detecting Data Contamination in RL Post-Training}

\author{
  Minju Gwak$^{1}$ \quad
  Minseo Kwak$^{1}$ \quad
  Dongseok Lee$^{1}$ \\
  \textbf{Guijin Son$^{2}$ \quad
  Alan Ritter$^{3}$ \quad
  Jaehyung Kim$^{1}$} \\
  $^{1}$Yonsei University \quad
  $^{2}$Seoul National University \quad
  $^{3}$Georgia Institute of Technology \\
  \texttt{mjgwak@yonsei.ac.kr, jaehyungk@yonsei.ac.kr}
}

\begin{document}
\maketitle
\begin{abstract}
Reinforcement learning (RL) post-training has shown to improve reasoning in large language models (LLMs). However, there has been little exploration on the problem of data contamination in RL post-training, potentially undermining generalization and evaluation reliability of the training process itself. Existing detection methods primarily rely on output-level signals such as likelihood or entropy, which become unreliable for RL-trained models since RL shapes behavior through trajectory-level rewards rather than token likelihoods. We propose LaRA, a layer-wise representation analysis framework for detecting contamination in RL post-trained LLMs. LaRA introduces three complementary metrics, measuring perturbation sensitivity, directional collapse, and local representation rigidity under controlled perturbations. We find that contamination produces progressive geometric deviations across layers, including amplified perturbation sensitivity, stronger directional collapse, and enhanced local rigidity. Based on our findings, we also develop a contamination detection protocol that aggregates representation-level deviations across layers and metrics. Experiments on RL-trained reasoning models show that our protocol outperforms existing output-level baselines for contamination detection.\footnote{The code is available at \url{https://github.com/talzoomanzoo/lara}.}
\end{abstract}

\section{Introduction}
Reinforcement learning (RL) has shown its effectiveness in training Large Language Models (LLMs) for complex reasoning tasks~\citep{guo2025deepseek, guha2025openthoughts, li2025limr, hochlehnert2025sober}.
However, it also raises a critical but underexplored issue of \textit{data contamination} in RL post-training~\citep{tao2025detecting, wang2025fragility, wu2026reasoning}, the inclusion of evaluation or benchmark samples within the RL training data. 
Contaminated samples can induce reward-driven overfitting and implicit memorization, undermining generalization and evaluation reliability.

\begin{figure}[t]
    \centering
    \includegraphics[width=\columnwidth]{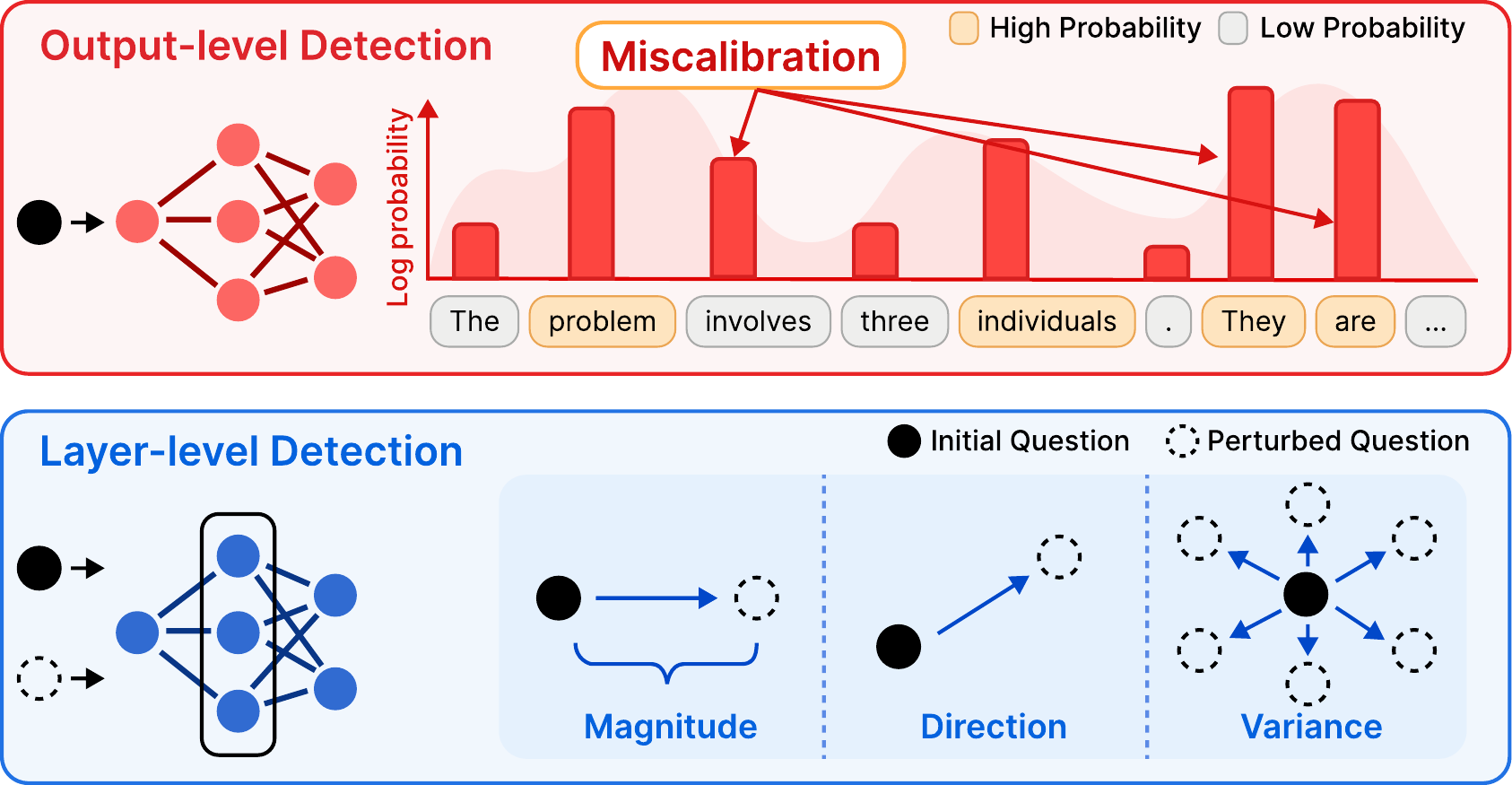}
    \caption{\textbf{Output vs. Representation-level detection.} Output-level signals are sensitive to output miscalibration, such as overconfidence of superficially plausible tokens. In contrast, representation geometry provides more robust and interpretable contamination signals.}
    \label{fig:main}
    \vspace{-1.5em}
\end{figure}

Prior work on data contamination in LLMs has mostly focused on pre-training or supervised fine-tuning (SFT) stages~\citep{zhang2024min, shi2023detecting, xie2024recall}, where memorization is typically characterized by higher token likelihoods or lower entropy~\citep{gonen2023demystifying}. 
Consequently, existing approaches primarily rely on output-level signals derived from model likelihoods or generation statistics. Recent work has extended this paradigm to reasoning trajectories for detecting data contamination in RL, using entropy or behavioral divergence across generation stages as contamination signals~\citep{tao2025detecting}. 

However, such output-level statistics can be unreliable due to the poor calibration of LLM output distributions, as shown in Figure~\ref{fig:main} \citep{leng2025taming, xiao2025restoring}.
Moreover, unlike pre-training or SFT, RL optimizes expected reward over entire reasoning trajectories rather than token-wise likelihoods, making likelihood-based behavioral signals less directly aligned with the underlying training objective. 
These challenges motivate a shift toward representation-level analysis, where memorization can be probed directly in the model's internal geometry, bypassing the calibration issues and objective mismatch that confound output-level signals.

We propose \textbf{LaRA}, a \textbf{La}yerwise \textbf{R}epresentation \textbf{A}nalysis framework for detecting data contamination in RL post-training. 
{Our key hypothesis is that RL-induced memorization produces abnormal representation responses under controlled perturbations: memorized samples become overly stable to semantically equivalent variations, yet exhibit disproportionately large representation shifts when memorized information is removed.} 
To test this, we construct structural control groups of semantically similar questions, apply consistent information masking, and analyze layer-wise representation dynamics across perturbations.

{
Specifically, our metrics are motivated by three complementary questions about memorized representations. First, does removing key information perturb memorized representations more strongly than non-memorized ones? To examine this question \textit{Representation Shift Magnitude (RSM)} measures how strongly representations change when important information is removed, capturing perturbation sensitivity.
Second, do the resulting changes become concentrated along shared dominant directions?
\textit{Directional Collapse (DC)} measures whether representation changes collapse toward shared dominant directions, indicating reduced representational diversity. 
Third, do memorized representations become locally rigid across semantically equivalent variations? \textit{Representation Stability Index (RSI)} quantifies how invariant representations remain across semantically similar variants, capturing local rigidity under meaning-preserving perturbations. 
Together, these metrics characterize distinct geometric signatures of RL-induced memorization.
}

Across multiple RL-trained models, we empirically show that contaminated samples exhibit consistent geometric abnormalities compared to non-trained samples. 
In particular, contaminated samples exhibit abnormal directional collapse, higher local representational rigidity, and greater sensitivity to information removal. 
Furthermore, our LaRA-based contamination detection score consistently outperforms output-level baselines, suggesting that representation geometry provides a more reliable signal of RL-induced memorization. 

\noindent In summary, our contributions are as follows:
\begin{itemize}[leftmargin=*, itemsep=0pt, topsep=2pt]
\item[$\circ$] We are the first to propose a representation-level framework as well as a training and evaluation setup for detecting contamination in RL post-training via stiffness and rigidity.
\item[$\circ$] We introduce a contamination-detection protocol that consistently outperforms output-level baselines across RL-trained models, achieving up to +9.6\% AUC improvement and 3.5$\times$ higher TPR@FPR=5\% compared to the strongest prior output-level method.
\item[$\circ$] We provide empirical insights into how RL training affects representation geometry across layers.
\end{itemize}
\raggedbottom
\section{Related Work}
\paragraph{Data Contamination.} 
Data contamination detection~\citep{golchin2024time, golchin2025data, deng2024investigating} is commonly formulated as a membership inference attack (MIA) problem~\citep{wu2025membership}, where contaminated samples are identified through behavioral differences between training and non-training data. 
Existing methods primarily exploit output-level statistics ~\citep{gonen2023demystifying, xie2024recall, zhang2024min, shi2023detecting, kwak2026gap, tao2025detecting}. 
While these signals are strong indicators of memorization under likelihood-maximization training (\textit{i.e.}, pre-training and SFT), they become unreliable for RL-trained models, since RL optimizes models through reward-driven exploration of reasoning trajectories rather than token-level likelihoods.
Contamination detection specifically for RL post-training, however, remains underexplored: existing attempts largely transfer the same output-level signals, \textit{e.g.}, entropy-based detection~\citep{tao2025detecting}. 
Consequently, they inherit the limitations above, often compounded by exploration dynamics.

\paragraph{Representation Dynamics in LLMs.} 
Recent work has increasingly leveraged representation dynamics in LLMs to study behaviors beyond outputs~\citep{kang2025scalable,lee2025training,gwak2025revisiting,zhao2025learning}. 
One line of work analyzes internal states and their evolution across layers to characterize properties emerging during post-training~\citep{bi2026reasoning,wang2024latent,hao2024training,li2025tracing}. 
Another line shows that semantic and behavioral attributes are encoded in hidden representations, where specific directions can be exploited to steer, detect, or modulate model behavior~\citep{turner2023steering,lee2024programming,li2023inference,roh2026embracing,wurgaft2026manifoldsteeringrevealsshared}. 
Closer to our setting, internal representations have also been used for contamination analysis: 
Kernel Divergence Score ~\citep{choi2025contaminated} quantifies contamination by measuring how fine-tuning on a benchmark dataset changes the similarity structure of sample embeddings. 
Neural Breadcrumbs~\citep{makhija2025neuralbreadcrumbsmembershipinference} extracts features from hidden states and attention patterns, while G-Drift MIA~\citep{ranjan2026gdriftmiamembershipinference} measures representation changes induced by a targeted gradient update.
Both methods demonstrate that internal model states contain informative membership signals, but train supervised classifiers using labeled member and non-member examples.
In contrast, LaRA directly derives an instance-level contamination score from layer-wise representation geometry without training an auxiliary classifier or updating the target model.
This training-free design isolates the informativeness of the representation signal itself and enables direct application to RL post-trained LLMs.
\section{LaRA: Layer-wise Representation Analysis to Detect RL Contamination}
\begin{figure*}[t]
    \centering
    \includegraphics[width=\textwidth]{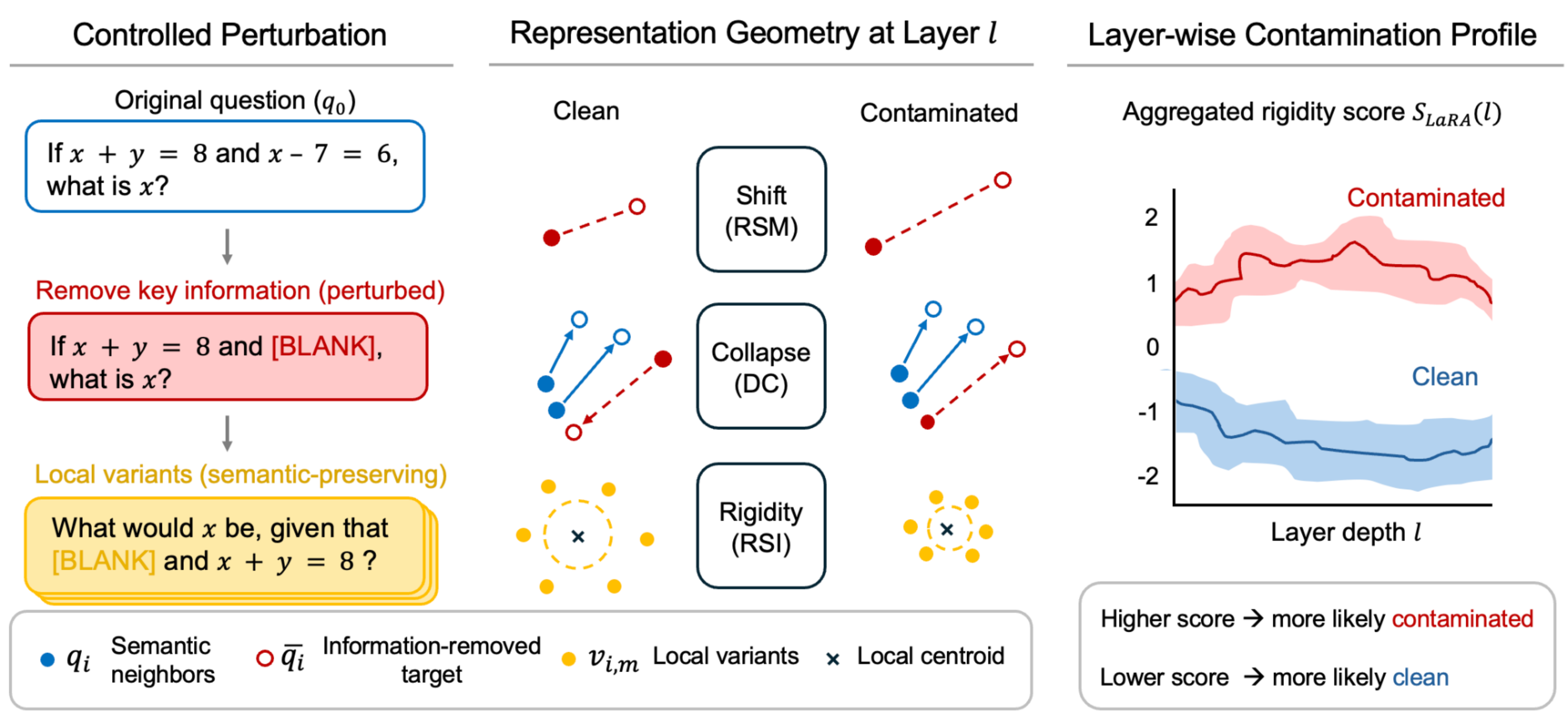}
    \caption{\textbf{Overview of LaRA, the proposed layer-wise representation geometry analysis framework.} 
    Given an original question $q_0$, we generate semantically similar questions and remove shared key information to construct perturbed variants. 
    Hidden representations are extracted across all transformer layers for original, perturbed, and paraphrased inputs. 
    We then compute three complementary geometric metrics: Representation Shift Magnitude (RSM), Directional Collapse (DC), and Representation Stability Index (RSI), which characterize perturbation sensitivity, directional organization, and local representation variability under controlled perturbations.}
    \label{fig:main}
\end{figure*}
We frame the problem of detecting data contamination during RL post-training as {Membership Inference Attack (MIA)}. 
Given an RL-trained model $\mathcal{M}$ and a candidate sample $x$, our goal is to determine membership $\mathcal{F}(M,x) \in \{0,1\}$, where $\mathcal{F}(M,x)=1$ indicates that $x$ was a member of the training dataset and therefore indicates contamination, while 0 indicates otherwise.  
The central question motivating our analyses is: \textit{do layer-wise representation signals behave differently between member and non-member samples?} 
To answer this, we introduce three complementary metrics.

\subsection{Contamination Dataset Construction}
To explore MIA in RL training setting, we construct controlled contamination benchmarks that support a two-stage analysis: (i) detecting contamination in released open-source RL checkpoints based on their known training data, and (ii) tracking how detection signals evolve under additional RL training that we perform on a controlled corpus. 

\paragraph{Evaluation set.} We construct a {contamination evaluation set} from the publicly open dataset of the three open-source RL-trained models (\texttt{EURUS-2-7B-PRIME} (Eurus)~\citep{cui2025process}, \texttt{LIMR}~\citep{li2025limr}, and \texttt{Olmo-3.1-7B-RL-Zero-Math} (Olmo) ~\citep{olmo2025olmo}).
For each model, we sample 30 Olympiad-level mathematics problems from its own RL training set as members, and pair them with 30 problems from AIME 2026~\citep{balunovic_srimatharena_2025} as non-members. This yields a balanced 60-sample evaluation set per model. Non-member split is shared across all three models, while the member split is model-specific.

\paragraph{Training set.} 
To study how contamination signals vary during continued RL training, we re-use each model's 30 member samples as deliberate contamination targets and augment them with 970 Olympiad-level problems drawn from the RL-MIA~\citep{tao2025detecting} Math dataset, yielding a 1{,}000-sample training corpus per model. 
Using this data, we resume RL training on each open-source checkpoint and track how member vs. non-member signals diverge during RL post-training. 
Further details on datasets are provided in Appendix~\ref{app:contam_data}.

\subsection{Three Metrics for Analysis}\label{sec:three-metrics}

\paragraph{Metric 1: Representation Shift Magnitude.}
To quantify how strongly a model's internal representation responds to the removal of important information, we introduce Representation Shift Magnitude (RSM). 
Given an original question $q_0$, we construct a set of semantically similar questions
\[
\mathcal{Q} = \{q_0, q_1, \dots, q_K\},
\]
where $K$ denotes the number of generated semantic neighbors excluding the original question. 
For each question $q_i \in \mathcal{Q}$, we apply an importance-based blanking operator $\textsc{BlankImportant}$ that removes key information spans while preserving the overall question structure:
\[
q_i^- \leftarrow \textsc{BlankImportant}(q_i, k),
\]
where $k$ denotes the number of inserted \texttt{[BLANK]} tokens.
Refer to Appendix~\ref{app:similar_question_prompt} for details of the perturbation construction process. Let $h_\ell(\cdot)$ denote the mean-pooled hidden representation extracted from transformer layer $\ell$, where $\ell \in \mathcal{L} = \{0,1,\dots,L-1\}$. For each $\ell$, we extract hidden representations $u_i = h_\ell(q_i)$ and $w_i = h_\ell(q_i^-)$, 
where $u_i, w_i \in \mathbb{R}^d$ and $d$ is the hidden representation dimension. 
We then compute the perturbation-induced representation shift $\Delta_i$ and define its magnitude $S_i$ as:
\[
S_i = \|\Delta_i\|_2, \quad \Delta_i = u_i - w_i
\]
where $\|\cdot\|_2$ denotes the Euclidean norm. To capture how anomalously the original question responds to perturbation relative to its semantic neighbors, we standardize its shift magnitude using the mean and standard deviation of the similar-question set:
\[
RSM_\ell
=
\frac{
S_0 - \mu_S
}{
\sigma_S + \epsilon
},
\]
where
\[
\mu_S
=
\frac{1}{K}
\sum_{i=1}^{K} S_i,~
\sigma_S
=
\sqrt{
\frac{1}{K-1}
\sum_{i=1}^{K}
(S_i-\mu_S)^2
},
\]
and $\epsilon > 0$ is a numerical stability constant. 
A high $RSM_\ell$ indicates that the original question exhibits a  larger representation shift under information removal compared to semantically similar questions, suggesting stronger perturbation sensitivity from memorization and risk of contamination.

\paragraph{Metric 2: Directional Collapse.}
We introduce Directional Collapse (DC) to characterize the directional organization of perturbation-induced representation changes. 
We first compute the average perturbation direction across similar questions:
\[
\bar{s}_\ell
=
\frac{1}{K}
\sum_{i=1}^{K}
\Delta_i,
\]
where $\bar{s}_\ell \in \mathbb{R}^d$ represents the average perturbation direction shared across the semantic group. 
DC is then defined as:
\[
DC_\ell
=
\frac{
\Delta_0^\top \bar{s}_\ell
}{
(\|\Delta_0\|_2+\epsilon)
(\|\bar{s}_\ell\|_2+\epsilon)
}.
\]
This quantity measures the cosine alignment between the original perturbation direction and the average perturbation direction of semantically similar questions. 
High $DC_\ell$ values indicate that perturbation responses are strongly aligned along a shared low-dimensional direction, whereas lower values indicate more distributed or heterogeneous perturbation dynamics.

\begin{figure*}[t]
    \centering

    \includegraphics[width=\textwidth]{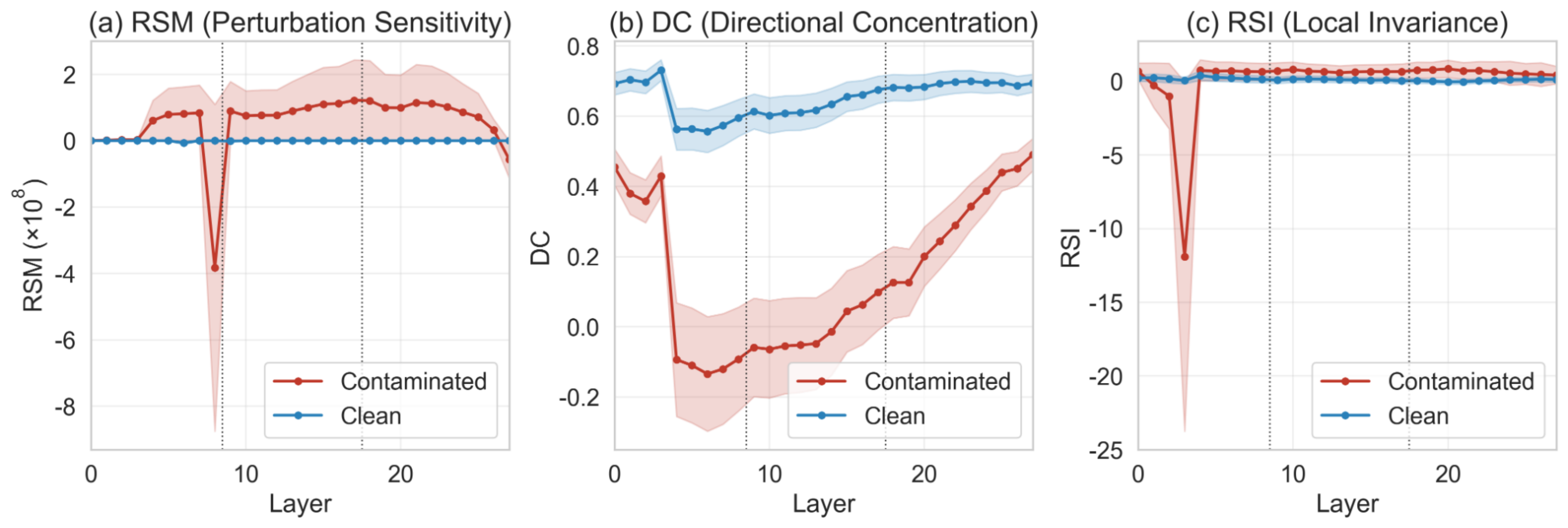}

    \caption{\textbf{Layer-wise representation geometry patterns of RL post-trained model.} We compare layer-wise representation geometry between contaminated and clean samples under input perturbations. Contaminated samples consistently exhibit deviating perturbation sensitivity (\textbf{RSM}), abnormal directional concentration dynamics (\textbf{DC}), and altered local representation variability patterns (\textbf{RSI}) across transformer layers, indicating that memorized samples form distinct and less robust internal representation structures compared to clean samples.}
    \label{fig:combined_geometry}
\end{figure*}

\paragraph{Metric 3: Representation Stability Index.}
Finally, we measure local representation stability under semantically preserving perturbations through the Representation Stability Index (RSI). 
For each perturbed question $q_i^-$, we generate $M$ paraphrastic variants while preserving the blank positions:
\[
\{v_{i,1}, \dots, v_{i,M}\}
\sim
\textsc{VariantGen}(q_i^-).
\]
We then extract their hidden representations:
\[
\phi_{i,m} = h_\ell(v_{i,m}),
\]
where $\phi_{i,m} \in \mathbb{R}^d$. 
Next, we compute the local representation centroid:
\[
\bar{\phi}_i
=
\frac{1}{M}
\sum_{m=1}^{M}
\phi_{i,m},
\]
and define its average deviation:
\[
R_i
=
\frac{1}{M}
\sum_{m=1}^{M}
\|
\phi_{i,m}-\bar{\phi}_i
\|_2.
\]
We then standardize the original question's local variability relative to its semantic neighbors:
\[
RSI_\ell
=
\frac{
R_0-\mu_R
}{
\sigma_R+\epsilon
},
\]
where
\[
\mu_R
=
\frac{1}{K}
\sum_{i=1}^{K} R_i,~ 
\sigma_R
=
\sqrt{
\frac{1}{K-1}
\sum_{i=1}^{K}
(R_i-\mu_R)^2
}.
\]

A high $RSI_\ell$ indicates that the original question exhibits larger local representation variability relative to semantically similar questions under paraphrastic perturbations, while lower values indicate more locally stable representation behavior.

\subsection{Layer-wise Analysis with Three Metrics}\label{sec:sec2-3}
Figure~\ref{fig:combined_geometry} shows the representation geometry patterns measured by the three metrics in Section~\ref{sec:three-metrics}. Contaminated samples consistently exhibit larger perturbation-induced representation shifts (RSM) than clean samples across most layers, while clean samples remain near zero throughout depth. In particular, contaminated samples sharply deviate around layers 7–9, indicating substantially higher sensitivity to targeted information removal and stronger dependence on memorized information. DC results further show that contaminated samples exhibit distinct directional concentration dynamics compared to clean samples. RSI results show that contaminated samples exhibit lower local representation variability, particularly in early layers, indicating more rigid and invariant local representation geometry under paraphrastic perturbations. Additional results are provided in Appendix~\ref{app:figure_geometry}.


\section{Contamination Detection Protocol}\label{sec:contamination_detection_protocol}

Motivated by Section~\ref{sec:sec2-3}, we formulate contamination detection as a layer-aware representation anomaly detection problem. 
We find that contaminated samples exhibit distinct representation profiles across depth, including amplified perturbation sensitivity, abnormal directional concentration dynamics, and local variability under controlled perturbations. 
Consequently, contamination should be characterized through \emph{deviation from clean geometric profiles} across multiple metrics and layers, rather than from isolated layer-wise statistics.

\paragraph{Step 1: Clean-reference Robust Standardization.}
Let $\mathcal{M} = \{\mathrm{RSM}, \mathrm{DC}, \mathrm{RSI}\}$ denote the set of representation geometry metrics, $\mathcal{L}$ denote the set of probed transformer layers, and $m_\ell(x)$ denote the value of metric $m$ at layer $\ell$ for sample $x$. 
The three metrics span several orders of magnitude in raw form, so we first apply a sign-preserving compression to tame their heavy-tailed regime while leaving values near zero unchanged:
\begin{equation}
\tilde{m}_\ell(x)
=
\operatorname{sign}\!\bigl(m_\ell(x)\bigr)\,
\log\!\bigl(1 + |m_\ell(x)|\bigr).
\end{equation}

For each $(m,\ell)$, we estimate the clean reference \emph{center} and \emph{scale} from non-contaminated validation samples $\mathcal{D}^{\mathrm{clean}}$:
\begin{align}
\mu^{\mathrm{clean}}_{m,\ell}
&=
\operatorname{median} \!\Bigl(
\tilde{m}_\ell(x) : x \in \mathcal{D}^{\mathrm{clean}}
\Bigr), \\
\sigma^{\mathrm{clean}}_{m,\ell}
&=
1.4826 \cdot
\operatorname{MAD} \!\Bigl(
\tilde{m}_\ell(x) : x \in \mathcal{D}^{\mathrm{clean}}
\Bigr),
\end{align}
where the factor $1.4826$ is the standard scaling to make median absolute deviation (MAD) a consistent estimator of the standard deviation under Gaussian noise (see Appendix~\ref{app:standard-scaling-factor}). 
The standardized geometric deviation of sample $x$ at $(m,\ell)$ is then:
\begin{equation}
z_{m,\ell}(x)
=
\frac{
\tilde{m}_\ell(x)
-
\mu^{\mathrm{clean}}_{m,\ell}
}{
\sigma^{\mathrm{clean}}_{m,\ell} + \epsilon
},
\end{equation}
with $\epsilon$ a small numerical constant. 
This formulation preserves the relative magnitude of geometric deviations while preventing a small number of extreme contaminated samples from inflating the clean-reference scale and washing out the signal for the rest of the population.

\begin{table*}[t]
\centering
\small
\renewcommand{\arraystretch}{1.00}

\begin{minipage}[t]{0.38\textwidth}
\centering
\setlength{\tabcolsep}{3pt}

\resizebox{\textwidth}{!}{
\begin{tabular}{l|cc|cc|cc}
\toprule

\multirow{2}{*}{\textbf{Method}}
& \multicolumn{2}{c|}{\textbf{Eurus}}
& \multicolumn{2}{c|}{\textbf{LIMR}}
& \multicolumn{2}{c}{\textbf{OLMO}} \\

\cmidrule(lr){2-3}
\cmidrule(lr){4-5}
\cmidrule(lr){6-7}

& \textbf{AUC} & \textbf{TPR}
& \textbf{AUC} & \textbf{TPR}
& \textbf{AUC} & \textbf{TPR} \\

\midrule

Recall
& 0.59 & 0.07
& 0.54 & 0.07
& 0.42 & 0.03 \\

CDD
& 0.38 & 0.00
& 0.64 & 0.00
& \underline{0.50} & 0.00 \\

Min-K\%
& 0.32 & 0.07
& 0.26 & 0.03
& \underline{0.50} & 0.03 \\

Min-K\%++
& 0.29 & 0.03
& 0.60 & 0.07
& 0.46 & \underline{0.07} \\

PPL
& 0.68 & 0.23
& \underline{0.73} & \underline{0.13}
& 0.49 & 0.07 \\

SC
& \underline{0.70} & \underline{0.27}
& 0.44 & 0.00
& 0.42 & 0.03 \\

\cmidrule(lr){1-7}

\rowcolor{lightblue}
$S_{LaRA}$
& 0.63 & 0.19
& \textbf{0.80} & \textbf{0.46}
& \textbf{0.54} & 0.04 \\

\rowcolor{lightblue}
SC + $S_{LaRA}$
& \textbf{0.73} & \textbf{0.31}
& 0.47 & 0.04
& \underline{0.50} & \textbf{0.08} \\

\bottomrule
\end{tabular}
}

\vspace{2mm}
{\small \textbf{(a)} Initial checkpoint results across models.}

\label{tab:init_results}
\end{minipage}
\hfill
\begin{minipage}[t]{0.58\textwidth}
\centering
\setlength{\tabcolsep}{2.5pt}

\resizebox{\textwidth}{!}{
\begin{tabular}{llcc|cc|cc|cc|cc|cc}
\toprule

&
& \multicolumn{6}{c|}{\textbf{Eurus}}
& \multicolumn{6}{c}{\textbf{LIMR}} \\

\cmidrule(lr){3-8}
\cmidrule(lr){9-14}

&
& \multicolumn{2}{c|}{\textbf{Init}}
& \multicolumn{2}{c|}{\textbf{E1}}
& \multicolumn{2}{c|}{\textbf{E2}}
& \multicolumn{2}{c|}{\textbf{Init}}
& \multicolumn{2}{c|}{\textbf{E1}}
& \multicolumn{2}{c}{\textbf{E2}} \\

\cmidrule(lr){3-4}
\cmidrule(lr){5-6}
\cmidrule(lr){7-8}
\cmidrule(lr){9-10}
\cmidrule(lr){11-12}
\cmidrule(lr){13-14}

& \textbf{Method}
& \textbf{AUC} & \textbf{TPR}
& \textbf{AUC} & \textbf{TPR}
& \textbf{AUC} & \textbf{TPR}
& \textbf{AUC} & \textbf{TPR}
& \textbf{AUC} & \textbf{TPR}
& \textbf{AUC} & \textbf{TPR} \\

\midrule

\multirow{8}{*}{}

& Recall
& 0.59 & 0.07
& 0.56 & 0.13
& 0.52 & 0.00
& 0.54 & 0.07
& 0.47 & 0.03
& 0.47 & 0.00 \\

& CDD
& 0.38 & 0.00
& 0.38 & 0.00
& 0.38 & 0.00
& 0.64 & 0.00
& 0.61 & 0.00
& 0.59 & 0.00 \\

& Min-K\%
& 0.32 & 0.07
& 0.31 & 0.07
& 0.33 & 0.07
& 0.26 & 0.03
& 0.26 & 0.00
& 0.26 & 0.00 \\

& Min-K\%++
& 0.29 & 0.03
& 0.26 & 0.03
& 0.20 & 0.03
& 0.60 & 0.07
& 0.58 & 0.07
& 0.63 & 0.00 \\

& PPL
& 0.68 & 0.23
& \textbf{0.69} & \underline{0.23}
& 0.67 & 0.17
& \underline{0.73} & \underline{0.13}
& \textbf{0.74} & \underline{0.13}
& \underline{0.74} & \underline{0.17} \\

& SC
& \underline{0.70} & \underline{0.27}
& 0.58 & 0.07
& \underline{0.78} & \underline{0.30}
& 0.44 & 0.00
& 0.40 & 0.00
& 0.46 & 0.00 \\

\cmidrule(lr){2-14}

\rowcolor{lightblue}
& $S_{LaRA}$
& 0.63 & 0.19
& \underline{0.68} & 0.19
& 0.70 & 0.15
& \textbf{0.80} & \textbf{0.46}
& \underline{0.72} & \textbf{0.23}
& \textbf{0.81} & \textbf{0.20} \\

\rowcolor{lightblue}
& SC + $S_{LaRA}$
& \textbf{0.73} & \textbf{0.31}
& 0.65 & \textbf{0.35}
& \textbf{0.79} & \textbf{0.38}
& 0.47 & 0.04
& 0.46 & 0.04
& 0.54 & 0.04 \\

\bottomrule
\end{tabular}
}

\vspace{2mm}
{\small \textbf{(b)} Results across Eurus and LIMR training checkpoints.}

\label{tab:combined_results}
\end{minipage}


\caption{
\textbf{Membership inference performance across different RL checkpoints and model families.}
(a) compares initial checkpoints across Eurus, LIMR, and OLMO, while
(b) analyzes performance evolution during RL training for Eurus and LIMR.
TPR denotes TPR@FPR=5\%. Details are in ~\ref{app:main_table_confidence}.
}
\label{tab:main_combined}
\end{table*}

\paragraph{Step 2: Metric-specific Anomaly Alignment.}
Our analyses show that contamination affects each metric through a different geometric mechanism.
Contaminated samples tend to exhibit elevated perturbation sensitivity in $\mathrm{RSM}$, directional concentration shifts in $\mathrm{DC}$, and lower local representation variability in $\mathrm{RSI}$.
Importantly, the raw $\mathrm{RSI}$ metric introduced in Section~3 measures local representation variability, such that larger values indicate greater variability.
As shown in Section~3.3, contaminated samples generally have lower $\mathrm{RSI}$ values than clean samples, reflecting greater local representational rigidity.
Therefore, before aggregating the metrics, we align their directions so that a larger aligned score consistently indicates stronger evidence of contamination:
\begin{equation}
\hat{z}_{m,\ell}(x)
=
\begin{cases}
\phantom{-}z_{m,\ell}(x),
& m = \mathrm{RSM}, \\
\phantom{-}z_{m,\ell}(x),
& m = \mathrm{DC}, \\
-z_{m,\ell}(x),
& m = \mathrm{RSI}.
\end{cases}
\end{equation}
For $\mathrm{RSM}$ and $\mathrm{DC}$, the original sign is preserved because their positive standardized deviations are aligned with the corresponding contamination-associated patterns.
For $\mathrm{RSI}$, we negate the standardized score because lower raw $\mathrm{RSI}$ values indicate the increased local rigidity associated with contamination.
This sign alignment is applied only when constructing the aggregated detection score and does not change the definition or interpretation of the underlying $\mathrm{RSI}$ metric.

\paragraph{Step 3: Layer-wise Aggregation.}
We aggregate the aligned deviations $\hat{z}_{m,\ell}(x)$ across the metric set $\mathcal{M}$ and layer set $\mathcal{L}$ to obtain a single per-sample score, where larger values indicate stronger overall deviation from the clean geometric profile.
\begin{equation}
S_{\mathrm{LaRA}}(x)
=
\frac{1}{|\mathcal{M}|\,|\mathcal{L}|}
\sum_{m \in \mathcal{M}}
\sum_{\ell \in \mathcal{L}}
\hat{z}_{m,\ell}(x).
\end{equation}

Because all $(m,\ell)$ contributions are standardized onto the same robust z-scale before aggregation, abnormalities arising from different layers and metrics can be consistently compared and combined within $S_{\mathrm{LaRA}}(x)$. Accordingly, their combination is intended not merely to maximize detection performance, but to offer a richer analysis of representation geometry under RL-induced memorization.

\section{Experiments}

\paragraph{Setups.}
We evaluate contamination detection performance using standard metrics for MIA~\citep{zhang2024min, tao2025detecting, kwak2026gap}. \textit{ROC-AUC} (AUC) measures the model’s ability to distinguish between member and non-member samples across all possible decision thresholds. \textit{TPR@FPR=5\%} reports the true positive rate (\textit{i.e.}, correctly identified members) when the false positive rate (\textit{i.e.}, non-members incorrectly flagged as members) is fixed at 5\%. We consider six representative baselines, Recall, CDD, Min-K\%, Min-K\%++, PPL, and Self-Critique (SC). Refer to Appendix~\ref{app:metrics_and_baselines} for further details.
\raggedbottom
\subsection{Main Results}
Table~\ref{tab:main_combined} shows that our proposed representation-based membership score, $S_{\mathrm{LaRA}}$, consistently achieves strong and stable detection performance across different RL model families and training checkpoints. In the initial checkpoints, $S_{\mathrm{LaRA}}$ attains the best overall performance on LIMR with an AUC of $0.80$ and TPR@FPR=5\% of $0.46$, substantially outperforming standard baselines such as Recall, Min-K\%, and SC. We also explore combining $S_{LaRA}$ with SC, the sota output-level detection method, to see the complementarity of the two detection regimes. Combining $S_{LaRA}$ with SC ($\mathrm{SC}+S_{\mathrm{LaRA}}$) achieves the strongest overall performance on Eurus, reaching an AUC of $0.73$ and TPR@FPR=5\% of $0.31$ at initialization, while also maintaining competitive performance throughout RL training. 
Across Eurus checkpoints, the combined score steadily improves from $(0.73, 0.31)$ to $(0.79, 0.38)$ in terms of (AUC, TPR@FPR=5\%), suggesting that representation-level contamination signals become increasingly separable during RL optimization. 
LIMR exhibits a similar trend for $S_{\mathrm{LaRA}}$, where performance remains consistently high across checkpoints, peaking at $(0.81, 0.20)$ at epoch2.
Although PPL occasionally shows relatively high AUC values, their TPR@FPR=5\% remains substantially lower and less stable than the proposed methods. PPL often relies on superficial token likelihood differences that can fluctuate across model families and RL stages, whereas $S_{\mathrm{LaRA}}$ and $\mathrm{SC}+S_{\mathrm{LaRA}}$ capture deeper geometric inconsistencies in hidden representations. Therefore, our approach leads to more reliable detection under strict low-FPR operating regimes critical for realistic settings.

\begin{table}[t]
\centering
\scriptsize
\renewcommand{\arraystretch}{1.08}
\setlength{\tabcolsep}{4pt}

\resizebox{\columnwidth}{!}{
\begin{tabular}{
ccc
|cc cc cc
}
\toprule

\textbf{RSM}
& \textbf{DC}
& \textbf{RSI}

& \multicolumn{2}{c|}{\textbf{Init}}
& \multicolumn{2}{c|}{\textbf{E1}}
& \multicolumn{2}{c}{\textbf{E2}} \\

\cmidrule(lr){4-5}
\cmidrule(lr){6-7}
\cmidrule(lr){8-9}

&
&
&
\textbf{AUC} & \textbf{TPR}
& \textbf{AUC} & \textbf{TPR}
& \textbf{AUC} & \textbf{TPR} \\

\midrule

{\color{red}\ding{55}}
& {\color{red}\ding{55}}
& {\color{green!60!black}\ding{51}}
& 0.60 & 0.19 & 0.51 & 0.15 & 0.56 & 0.16 \\

{\color{red}\ding{55}}
& {\color{green!60!black}\ding{51}}
& {\color{red}\ding{55}}
& 0.71 & 0.42 & 0.66 & 0.08 & 0.76 & 0.24 \\

{\color{green!60!black}\ding{51}}
& {\color{red}\ding{55}}
& {\color{red}\ding{55}}
& 0.75 & 0.12 & 0.69 & 0.08 & 0.71 & 0.04 \\

\midrule

{\color{red}\ding{55}}
& {\color{green!60!black}\ding{51}}
& {\color{green!60!black}\ding{51}}
& 0.76 & 0.46 & 0.69 & 0.15 & 0.81 & 0.32 \\

{\color{green!60!black}\ding{51}}
& {\color{red}\ding{55}}
& {\color{green!60!black}\ding{51}}
& 0.74 & 0.27 & 0.68 & 0.08 & 0.71 & 0.12 \\

{\color{green!60!black}\ding{51}}
& {\color{green!60!black}\ding{51}}
& {\color{red}\ding{55}}
& 0.78 & 0.31 & 0.71 & 0.19 & 0.79 & 0.12 \\

\midrule

\rowcolor{lightblue}
{\color{green!60!black}\ding{51}}
& {\color{green!60!black}\ding{51}}
& {\color{green!60!black}\ding{51}}
& 0.80 & 0.46 & 0.72 & 0.23 & 0.81 & 0.20 \\

\bottomrule
\end{tabular}
}

\caption{
\textbf{Metric ablations across RL training epochs.}
}
\vspace{-1.0em}
\label{tab:metric_ablation_limr}
\end{table}

\subsection{Additional Analyses}
\paragraph{Metric Ablations.}
Table~\ref{tab:metric_ablation_limr} shows that combining all three components (RSM, DC, and RSI) consistently achieves the best overall performance across RL checkpoints, improving both AUC and robustness to training-stage shifts. While \textbf{DC} provides the strongest standalone discrimination signal, its performance varies more across epochs, particularly in TPR@FPR=5\%, indicating reduced robustness when used alone. In contrast, \textbf{RSM} and \textbf{RSI} individually produce weaker detection performance but contribute to improving generalization under RL post-training. Removing an individual component from the full metric consistently degrades performance, showing that the final score benefits from jointly modeling perturbation sensitivity, directional representation geometry, and local invariance. Overall, the results suggest that robust contamination detection requires integrating multiple representation-level signals rather than relying on a single geometric statistic.
\setlength{\dbltextfloatsep}{4pt}

\begin{table*}[t]
\vspace{-0.1in}
\centering
\small
\renewcommand{\arraystretch}{0.9}

\resizebox{\textwidth}{!}{
\begin{tabular}{
p{2.3cm}
p{11.3cm}
p{0.9cm}
p{1.0cm}
p{1.0cm}
p{1.0cm}
}
\toprule

\textbf{data\_source} 
& \textbf{prompt} 
& \textbf{member} 
& \textbf{RSM} 
& \textbf{DC} 
& \textbf{RSI} \\

\midrule

\texttt{numina\_amc\_aime}
&
Let the set $S = \{P_1, P_2, \dots, P_{12}\}$ consist of the twelve vertices of a regular $12$-gon. 
A subset $Q$ of $S$ is called ``communal'' if there is a circle such that all points of $Q$ are inside the circle, and all points of $S$ not in $Q$ are outside of the circle. 
How many communal subsets are there? (Note that the empty set is a communal subset.)

Present the answer in LaTeX format: $\backslash$boxed\{Your answer\}
&
1
&
0.151
&
0.423
&
0.310 \\

\midrule

\texttt{aime26}
&
Let triangle ABC have side lengths AB = 13, BC = 14, and CA = 15. Triangle A'B'C' is obtained by rotating triangle ABC about its circumcenter so that AC is perpendicular BC, with A' and B not on the same side of line B'C'. Find the integer closest to the area of hexagon AA'CC'BB'.Present the answer in LaTeX format: $\backslash$boxed\{Your answer\}
&
0
&
1.191
&
9.765
&
0.370 \\

\bottomrule
\end{tabular}
}

\caption{
\textbf{Error analysis.} (1) member sample often undetected, and (2) non-member sample often wrongly detected.
}
\vspace{0.2em}

\label{tab:error}

\end{table*}

\paragraph{Beta Sweep over $S_{LaRA}$ and SC Mix.}
We sweep the mixture weight $\beta$ in
$\mathrm{mix}
=
\beta\,(\text{SC})
+
(1-\beta)\,(S_{\mathrm{LaRA}})$
on the member-detection benchmark ($\beta\in\{0,0.25,0.5,0.65,0.75,1\}$). 
The optimal balance between SC and $S_{LaRA}$ is strongly model-dependent. 
For Eurus, performance improves as more weight is assigned to SC, with AUC peaking at the default $\beta=0.65$ and TPR@FPR=5\% at $\beta=0.75$. 
In contrast, LIMR performs best with only the $S_{LaRA}$, with performance degrading as $\beta$ increases. 
For OLMO, AUC is highest at $\beta=0$, while TPR@FPR=5\% peaks at $\beta=0.65$. 
Overall, these results suggest that no single mixture weight is universally optimal; however, despite not being tuned per model, the shared default $\beta=0.65$ still achieves competitive overall performance, consistent with the main results.


\begin{figure}
    \includegraphics[width=\columnwidth]{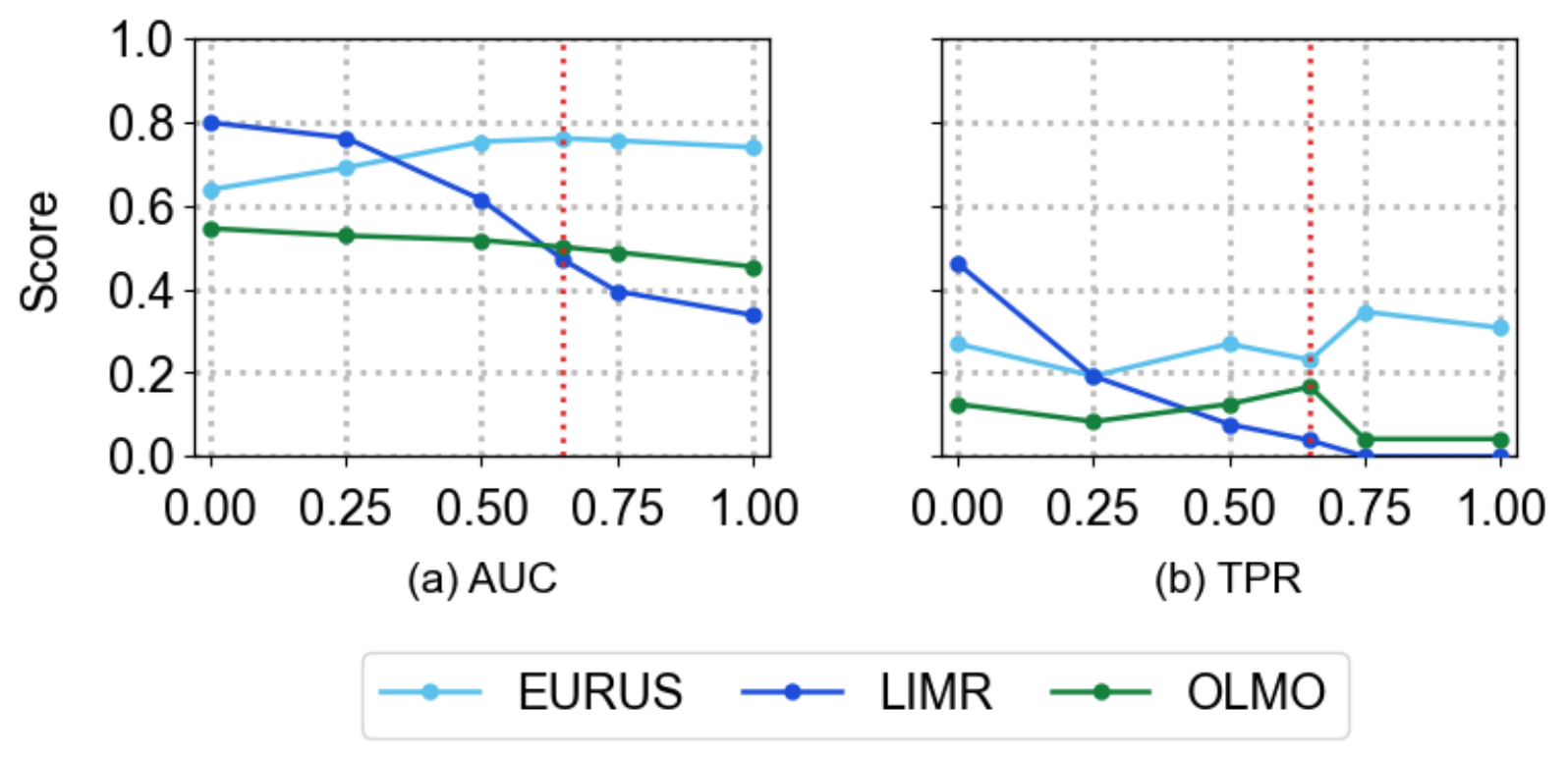}
    \caption{\textbf{Beta sweep over score mix.} Default $\beta$ = 0.65.}
    \label{fig:beta_sweep}
    \vspace{-0.5em}
\end{figure}

\begin{figure*}[t]
    \centering

    \begin{minipage}[t]{0.65\textwidth}
        \centering
        \includegraphics[width=\textwidth]{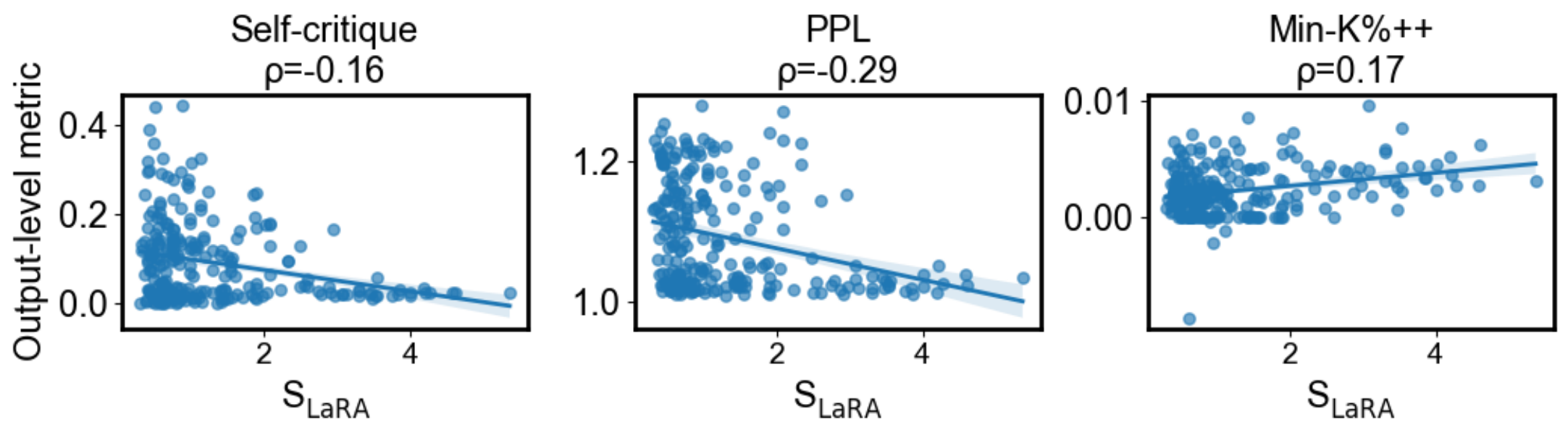}
    \end{minipage}
    \hfill
    \begin{minipage}[t]{0.30\textwidth}
        \centering
        \includegraphics[width=\textwidth]{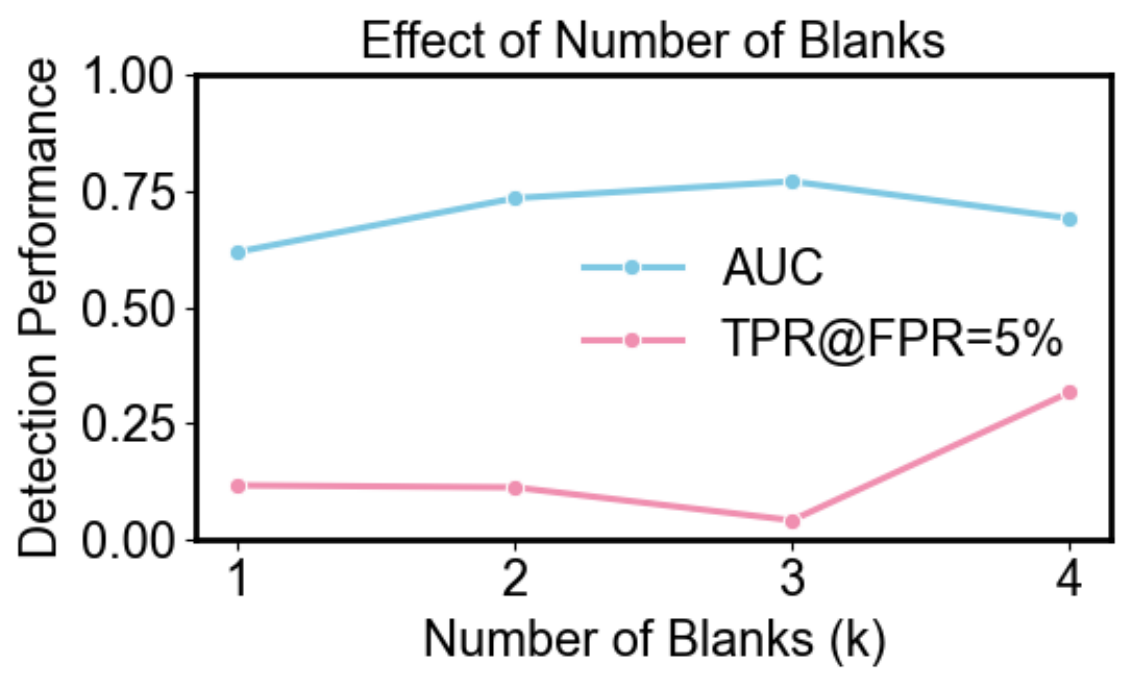}
    \end{minipage}

    \caption{\label{fig:perturb_set_size}
    \textbf{(a) Correlation plot of output-level signals and $S_{\mathrm{LaRA}}$}.
    \hfill
    \textbf{(b) Analysis on the \# of perturbations.}
    }
    \vspace{0.5em}

\end{figure*}

\paragraph{Correlation with Output-level Metrics.}
Figure~\ref{fig:perturb_set_size}(a) shows the correlation between the proposed $S_{\mathrm{LaRA}}$ and several output-level metrics.
Higher $S_{\mathrm{LaRA}}$ values are negatively correlated with SC ($\rho=-0.16$) and PPL ($\rho=-0.29$), while exhibiting a weak positive correlation with Min-K\%++ ($\rho=0.17$). 
Additionally, samples with low $S_{\mathrm{LaRA}}$ display substantially larger variability across all metrics, whereas high-$S_{\mathrm{LaRA}}$ samples are concentrated within narrower output regimes. Correlations suggest that stronger contamination-related geometric deviations are associated with increasingly confident, less reflective, and more behaviorally concentrated generations. 
In particular, high-$S_{\mathrm{LaRA}}$ samples occupy narrower output regimes characterized by reduced variability across output-level metrics.

\paragraph{Analysis on Number of Perturbations.} The perturbation-count sweep on Eurus (Figure~\ref{fig:perturb_set_size}(b)) shows that $S_{\mathrm{LaRA}}$ remains relatively stable across different numbers of blanks. Although AUC improves from the default $k=1$ (0.62) to a peak at $k=3$ (0.77), TPR@FPR=5\% only substantially increases at $k=4$ (0.32), where AUC slightly declines (0.69). 
This indicates that increasing the number of blanks can sometimes slightly strengthen detection, but the gains depend on the evaluation metric and operating regime. 
Overall, the method is robust to the choice of $k$, and the default setting $k=1$ already provides competitive performance without requiring additional perturbation variants.

\begin{figure}
    \centering
    \includegraphics[width=\columnwidth]{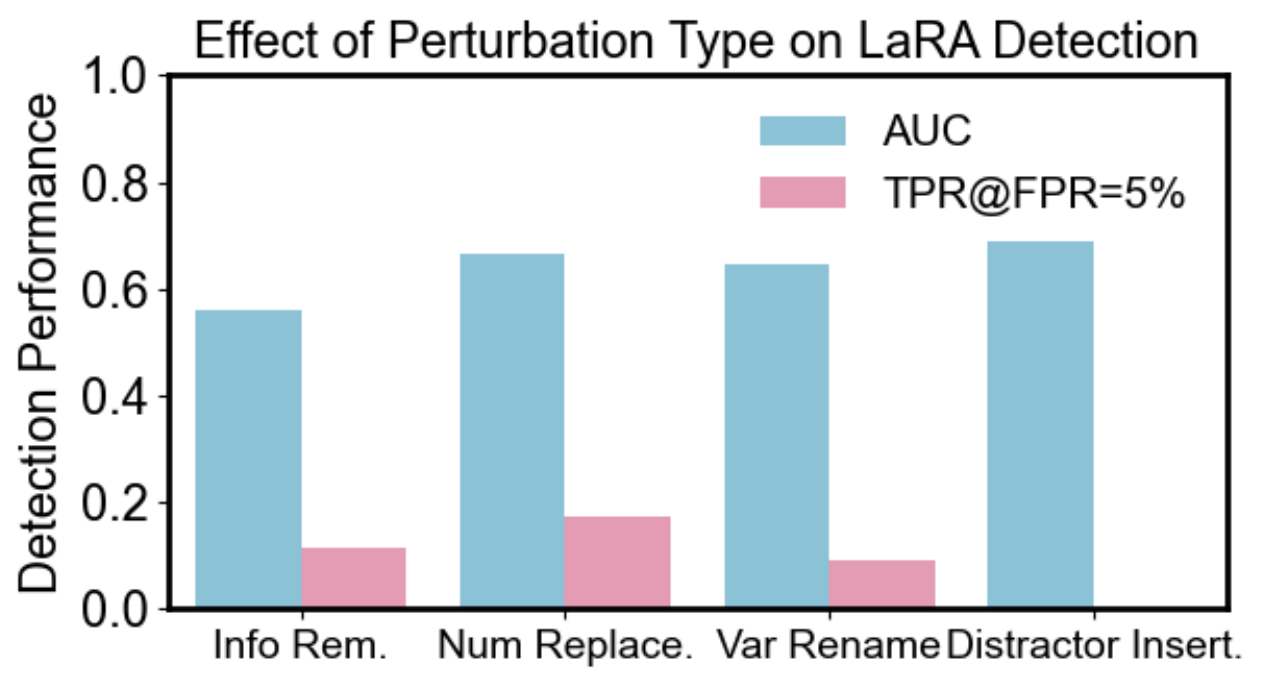}
    \caption{\textbf{Perturbation type analysis.} Results show that $S_{LaRA}$ is robust to perturbation types.}
    \label{fig:perturb_type}
    \vspace{-1.0em}
\end{figure}

\paragraph{Analysis on Perturbation Types.}
The perturbation analysis in Figure~\ref{fig:perturb_type} shows that LaRA remains relatively robust across different perturbation types, with all variants achieving comparable AUC values between 0.56 and 0.69. 
Although \texttt{Distractor Insert.} achieves the highest AUC (0.69) and \texttt{Num Replace.} obtains the best TPR@FPR=5\% (0.17), the default setting based on \texttt{Info Rem.}, the most naive method in making perturbations, still maintains reasonable detection performance without requiring semantically invasive modifications. 
In contrast to perturbations that directly alter numerical or semantic reasoning components, the default perturbation preserves the original problem structure more conservatively while still producing distinguishable contamination signals. 
Overall, results suggest that LaRA does not rely on a single perturbation type and exhibits stable detection behavior across diverse perturbation strategies.

\paragraph{Error analysis.}
Failure cases presented in Table~\ref{tab:error} suggest that $S_{\mathrm{LaRA}}$ misses member examples whose representation geometry remains close to the non-member manifold. The false-negative member sample exhibits uniformly low component scores, with RSM, DC, and RSI values of $0.151$, $0.423$, and $0.310$, respectively, producing a low aggregate score of $0.295$, well below the detection threshold. 
This indicates that the sample does not induce strong perturbation sensitivity, directional concentration, or local invariance, causing its layer-wise representation trajectory to resemble that of a clean example. Conversely, the non-member example that is falsely detected exhibits an abnormally high $S_{\mathrm{LaRA}}$ score, primarily driven by an unusually large DC value. Although the sample is not a member, its representation geometry deviates substantially from the typical clean distribution, leading the detector to classify it as contaminated.

\section{Conclusion}

We introduce LaRA, a layer-wise representation analysis framework for contamination detection in RL-trained models. 
Unlike prior methods, LaRA detects contamination through perturbation-induced representation geometry across layers, measured with three proposed metrics. 
Based on layer-wise analysis, we propose a detection framework that aggregates geometric deviations across metrics and layers. 
Experiments across various RL-trained models demonstrate that representation-level signals complement output-level methods and improve detection performance. 
Overall, our results show that contamination in RL-trained LLMs is strongly reflected in internal representation geometry, highlighting the effectiveness of representation-level auditing.

\section*{Limitations}

Despite its new findings and effectiveness, our work has several limitations.
First, LaRA requires access to intermediate hidden representations and is therefore currently limited to white-box settings. 
This requirement reflects the primary objective of our study: to investigate whether representation geometry provides a more reliable contamination signal than output-level statistics under RL post-training, rather than to develop a universally deployable detector.
Nevertheless, this constraint limits the applicability of LaRA to closed-source models accessible only through black-box APIs.
An important direction for future work is to translate our representation-level insights into practical black-box detectors, for example, by learning output-level proxies that approximate hidden-state geometry.

Second, LaRA relies on representation extraction under multiple semantic perturbations and layer-wise hidden-state analyses, introducing additional computational overhead compared with lightweight output-level approaches.
In particular, the framework requires generating perturbed variants, extracting intermediate representations across transformer layers, and aggregating multiple geometric statistics, making inference more expensive than methods operating solely on final outputs or token probabilities. 

Third, although the proposed framework consistently improves contamination detection over existing baselines, detection remains imperfect for challenging examples whose representation geometry closely overlaps with the clean distribution.
This suggests that some memorized samples may not produce sufficiently distinctive internal signatures for reliable separation. 
Moreover, while our analyses reveal consistent trends across models and training checkpoints, the precise relationship between RL post-training dynamics and representation-level memorization remains only partially understood. 
Future work may explore more computationally efficient perturbation strategies, stronger representation aggregation methods, and deeper investigations into the causal mechanisms underlying memorization and contamination in reasoning models.

\section*{Broader Impact and Ethical Implications}

This work contributes to improving transparency and reliability in the evaluation of RL-trained LLMs by introducing a representation-level framework for contamination detection. 
Stronger contamination auditing can help researchers better assess benchmark integrity, reduce hidden memorization effects, and improve the trustworthiness of reported reasoning capabilities. 
At the same time, contamination detection methods may potentially be adapted for membership inference or unauthorized dataset auditing, which could raise privacy or data governance concerns when applied to sensitive or proprietary data. 
Our experiments are conducted exclusively on publicly available open-source models and public mathematical datasets, and we do not use private or personally identifiable information. 
Overall, we believe representation-level auditing should be used responsibly as one component of broader efforts toward reliable, transparent, and reproducible evaluation of large language models.

\section*{Acknowledgments}

Minju Gwak, Minseo Kwak, Dongseok Lee, and Jaehyung Kim are affiliated with the Department of Artificial Intelligence at Yonsei University.
This research was supported in part by Institute for Information \& communications Technology Planning \& Evaluation (IITP) grant funded by the Korea government (MSIT) (No. RS-2020-II201361, Artificial Intelligence Graduate School Program (Yonsei University); No. RS-2025-02215344, Development of AI Technology with Robust and Flexible Resilience Against Risk Factors).
\bibliography{custom}

@article{guo2025deepseek,
  title={Deepseek-r1: Incentivizing reasoning capability in llms via reinforcement learning},
  author={Guo, Daya and Yang, Dejian and Zhang, Haowei and Song, Junxiao and Wang, Peiyi and Zhu, Qihao and Xu, Runxin and Zhang, Ruoyu and Ma, Shirong and Bi, Xiao and others},
  journal={arXiv preprint arXiv:2501.12948},
  year={2025}
}

@inproceedings{guha2025openthoughts,
  title={Openthoughts: Data recipes for reasoning models},
  author={Guha, Etash and Marten, Ryan and Keh, Sedrick and Raoof, Negin and Smyrnis, Georgios and Bansal, Hritik and Nezhurina, Marianna and Mercat, Jean and Vu, Trung and Sprague, Zayne and others},
  booktitle={International Conference on Learning Representations},
  year={2025}
}

@article{li2025limr,
  title={Limr: Less is more for rl scaling},
  author={Li, Xuefeng and Zou, Haoyang and Liu, Pengfei},
  journal={arXiv preprint arXiv:2502.11886},
  year={2025}
}

@inproceedings{tao2025detecting,
  title={Detecting data contamination from reinforcement learning post-training for large language models},
  author={Tao, Yongding and Wang, Tian and Dong, Yihong and Liu, Huanyu and Zhang, Kechi and Hu, Xiaolong and Li, Ge},
  booktitle={International Conference on Learning Representations},
  year={2025}
}

@inproceedings{wang2025fragility,
  title={On the fragility of benchmark contamination detection in reasoning models},
  author={Wang, Han and Li, Haoyu and Ko, Brian and Zhang, Huan},
  booktitle={International Conference on Learning Representations},
  year={2025}
}

@inproceedings{wu2025membership,
  title={Membership inference attacks on large-scale models: A survey},
  author={Wu, Hengyu and Cao, Yang},
  booktitle={Advances in Neural Information Processing Systems},
  year={2025}
}

@inproceedings{xie2024recall,
  title={Recall: Membership inference via relative conditional log-likelihoods},
  author={Xie, Roy and Wang, Junlin and Huang, Ruomin and Zhang, Minxing and Ge, Rong and Pei, Jian and Gong, Neil Zhenqiang and Dhingra, Bhuwan},
  booktitle={Proceedings of the 2024 Conference on Empirical Methods in Natural Language Processing},
  pages={8671--8689},
  year={2024}
}

@inproceedings{gonen2023demystifying,
  title={Demystifying prompts in language models via perplexity estimation},
  author={Gonen, Hila and Iyer, Srini and Blevins, Terra and Smith, Noah A and Zettlemoyer, Luke},
  booktitle={Findings of the Association for Computational Linguistics: EMNLP 2023},
  pages={10136--10148},
  year={2023}
}

@inproceedings{dong2024generalization,
  title={Generalization or memorization: Data contamination and trustworthy evaluation for large language models},
  author={Dong, Yihong and Jiang, Xue and Liu, Huanyu and Jin, Zhi and Gu, Bin and Yang, Mengfei and Li, Ge},
  booktitle={Findings of the Association for Computational Linguistics: ACL 2024},
  pages={12039--12050},
  year={2024}
}

@inproceedings{zhang2024min,
  title={Min-k\%++: Improved baseline for detecting pre-training data from large language models},
  author={Zhang, Jingyang and Sun, Jingwei and Yeats, Eric and Ouyang, Yang and Kuo, Martin and Zhang, Jianyi and Yang, Hao Frank and Li, Hai},
  booktitle={International Conference on Learning Representations},
  year={2024}
}

@inproceedings{shi2023detecting,
  title={Detecting pretraining data from large language models},
  author={Shi, Weijia and Ajith, Anirudh and Xia, Mengzhou and Huang, Yangsibo and Liu, Daogao and Blevins, Terra and Chen, Danqi and Zettlemoyer, Luke},
  booktitle={International Conference on Learning Representations},
  year={2023}
}

@misc{
bi2026reasoning,
title={Reasoning Self-Evaluation via Trajectory Dynamics Modeling},
author={Jinhe Bi and Danqi Yan and Yifan Wang and Wenke Huang and Haokun Chen and Guancheng Wan and Mang Ye and Xun Xiao and Hinrich Schuetze and Volker Tresp and Yunpu Ma},
year={2026},
}

@inproceedings{kang2025scalable,
  title={Scalable best-of-n selection for large language models via self-certainty},
  author={Kang, Zhewei and Zhao, Xuandong and Song, Dawn},
  booktitle={Advances in Neural Information Processing Systems},
  year={2025}
}

@article{gwak2025revisiting,
  title={Revisiting the Uniform Information Density Hypothesis in LLM Reasoning Traces},
  author={Gwak, Minju and Son, Guijin and Kim, Jaehyung},
  journal={arXiv preprint arXiv:2510.06953},
  year={2025}
}

@inproceedings{choi2025contaminated,
  title={How contaminated is your benchmark? quantifying dataset leakage in large language models with kernel divergence},
  author={Choi, Hyeong Kyu and Khanov, Maxim and Wei, Hongxin and Li, Yixuan},
  booktitle={International Conference on Machine Learning},
  year={2025}
}

@inproceedings{wu2026reasoning,
  title={Reasoning or memorization? unreliable results of reinforcement learning due to data contamination},
  author={Wu, Mingqi and Zhang, Zhihao and Dong, Qiaole and Xi, Zhiheng and Zhao, Jun and Jin, Senjie and Fan, Xiaoran and Zhou, Yuhao and Lv, Huijie and Zhang, Ming and others},
  booktitle={Proceedings of the AAAI Conference on Artificial Intelligence},
  volume={40},
  number={40},
  pages={33944--33952},
  year={2026}
}

@inproceedings{wang2024latent,
  title={Latent space chain-of-embedding enables output-free llm self-evaluation},
  author={Wang, Yiming and Zhang, Pei and Yang, Baosong and Wong, Derek F and Wang, Rui},
  booktitle={International Conference on Learning Representations},
  year={2024}
}

@inproceedings{zhao2025learning,
  title={Learning to reason without external rewards},
  author={Zhao, Xuandong and Kang, Zhewei and Feng, Aosong and Levine, Sergey and Song, Dawn},
  booktitle={Advances in Neural Information Processing Systems},
  year={2025}
}

@inproceedings{hao2024training,
  title={Training large language models to reason in a continuous latent space},
  author={Hao, Shibo and Sukhbaatar, Sainbayar and Su, DiJia and Li, Xian and Hu, Zhiting and Weston, Jason and Tian, Yuandong},
  booktitle={Conference on Language Modeling},
  year={2024}
}

@article{cui2025process,
  title={Process reinforcement through implicit rewards},
  author={Cui, Ganqu and Yuan, Lifan and Wang, Zefan and Wang, Hanbin and Zhang, Yuchen and Chen, Jiacheng and Li, Wendi and He, Bingxiang and Fan, Yuchen and Yu, Tianyu and others},
  journal={arXiv preprint arXiv:2502.01456},
  year={2025}
}

@inproceedings{hochlehnert2025sober,
  title={A sober look at progress in language model reasoning: Pitfalls and paths to reproducibility},
  author={Hochlehnert, Andreas and Bhatnagar, Hardik and Udandarao, Vishaal and Albanie, Samuel and Prabhu, Ameya and Bethge, Matthias},
  booktitle={Conference on Language Modeling},
  year={2025}
}

@misc{balunovic_srimatharena_2025,
  title = {MathArena: Evaluating LLMs on Uncontaminated Math Competitions},
  author = {Mislav Balunović and Jasper Dekoninck and Ivo Petrov and Nikola Jovanović and Martin Vechev},
  copyright = {MIT},
  publisher = {SRI Lab, ETH Zurich},
  month = feb,
  year = {2025},
}

@article{sheng2024hybridflow,
  title   = {HybridFlow: A Flexible and Efficient RLHF Framework},
  author  = {Guangming Sheng and Chi Zhang and Zilingfeng Ye and Xibin Wu and Wang Zhang and Ru Zhang and Yanghua Peng and Haibin Lin and Chuan Wu},
  year    = {2024},
  journal = {arXiv preprint arXiv: 2409.19256}
}

@article{kwak2026gap,
  title={Gap-K\%: Measuring Top-1 Prediction Gap for Detecting Pretraining Data},
  author={Kwak, Minseo and Kim, Jaehyung},
  journal={arXiv preprint arXiv:2601.19936},
  year={2026}
}

@article{li2025tracing,
  title={Tracing the representation geometry of language models from pretraining to post-training},
  author={Li, Melody Zixuan and Agrawal, Kumar Krishna and Ghosh, Arna and Teru, Komal Kumar and Santoro, Adam and Lajoie, Guillaume and Richards, Blake A},
  journal={arXiv preprint arXiv:2509.23024},
  year={2025}
}

@inproceedings{lee2024programming,
  title={Programming refusal with conditional activation steering},
  author={Lee, Bruce W and Padhi, Inkit and Ramamurthy, Karthikeyan Natesan and Miehling, Erik and Dognin, Pierre and Nagireddy, Manish and Dhurandhar, Amit},
  booktitle={International Conference on Learning Representations},
  year={2024}
}

@article{turner2023steering,
  title={Steering language models with activation engineering},
  author={Turner, Alexander Matt and Thiergart, Lisa and Leech, Gavin and Udell, David and Vazquez, Juan J and Mini, Ulisse and MacDiarmid, Monte},
  journal={arXiv preprint arXiv:2308.10248},
  year={2023}
}

@inproceedings{li2023inference,
  title={Inference-time intervention: Eliciting truthful answers from a language model},
  author={Li, Kenneth and Patel, Oam and Vi{\'e}gas, Fernanda and Pfister, Hanspeter and Wattenberg, Martin},
  booktitle={Advances in Neural Information Processing Systems},
}

@article{roh2026embracing,
  title={Embracing Anisotropy: Turning Massive Activations into Interpretable Control Knobs for Large Language Models},
  author={Roh, Youngji and Cho, Hyunjin and Kim, Jaehyung},
  journal={arXiv preprint arXiv:2603.00029},
  year={2026}
}

@misc{wurgaft2026manifoldsteeringrevealsshared,
      title={Manifold Steering Reveals the Shared Geometry of Neural Network Representation and Behavior}, 
      author={Daniel Wurgaft and Can Rager and Matthew Kowal and Vasudev Shyam and Sheridan Feucht and Usha Bhalla and Tal Haklay and Eric Bigelow and Raphael Sarfati and Thomas McGrath and Owen Lewis and Jack Merullo and Noah Goodman and Thomas Fel and Atticus Geiger and Ekdeep Singh Lubana},
      year={2026},
      eprint={2605.05115},
      archivePrefix={arXiv},
      primaryClass={cs.LG},
}

@inproceedings{kwon2023efficient,
  title={Efficient memory management for large language model serving with pagedattention},
  author={Kwon, Woosuk and Li, Zhuohan and Zhuang, Siyuan and Sheng, Ying and Zheng, Lianmin and Yu, Cody Hao and Gonzalez, Joseph and Zhang, Hao and Stoica, Ion},
  booktitle={Proceedings of the 29th symposium on operating systems principles},
  pages={611--626},
  year={2023}
}

@misc{openrouter,
  title = {OpenRouter: Unified API for {AI} Models},
  author = {OpenRouter},
  year = {2024},
}

@misc{openai2024gpt4omini,
  title={GPT-4o mini: advancing cost-efficient intelligence},
  author={{OpenAI}},
  year={2024},
}

@book{hampel1986robust,
  title={Robust Statistics: The Approach Based on Influence Functions},
  author={Hampel, Frank R. and Ronchetti, Elvezio M. and Rousseeuw, Peter J. and Stahel, Werner A.},
  year={1986},
  publisher={Wiley}
}

@book{huber2009robust,
  title={Robust Statistics},
  author={Huber, Peter J. and Ronchetti, Elvezio M.},
  edition={2},
  year={2009},
  publisher={Wiley}
}

@article{rousseeuw1993alternatives,
  title={Alternatives to the Median Absolute Deviation},
  author={Rousseeuw, Peter J. and Croux, Christophe},
  journal={Journal of the American Statistical Association},
  volume={88},
  number={424},
  pages={1273--1283},
  year={1993}
}

@book{rousseeuw1987robust,
  title={Robust Regression and Outlier Detection},
  author={Rousseeuw, Peter J. and Leroy, Annick M.},
  year={1987},
  publisher={Wiley}
}

@book{maronna2019robust,
  title={Robust Statistics: Theory and Methods (with R)},
  author={Maronna, Ricardo A. and Martin, R. Douglas and Yohai, Victor J. and Salibi{\'a}n-Barrera, Matias},
  edition={2},
  year={2019},
  publisher={Wiley}
}

@article{olmo2025olmo,
  title={Olmo 3},
  author={Olmo, Team and Ettinger, Allyson and Bertsch, Amanda and Kuehl, Bailey and Graham, David and Heineman, David and Groeneveld, Dirk and Brahman, Faeze and Timbers, Finbarr and Ivison, Hamish and others},
  journal={arXiv preprint arXiv:2512.13961},
  year={2025}
}

@inproceedings{golchin2024time,
  title={Time travel in llms: Tracing data contamination in large language models},
  author={Golchin, Shahriar and Surdeanu, Mihai},
  booktitle={International Conference on Learning Representations},
  volume={2024},
  pages={43008--43029},
  year={2024}
}

@article{golchin2025data,
  title={Data contamination quiz: A tool to detect and estimate contamination in large language models},
  author={Golchin, Shahriar and Surdeanu, Mihai},
  journal={Transactions of the Association for Computational Linguistics},
  volume={13},
  pages={809--830},
  year={2025},
  publisher={MIT Press 255 Main Street, 9th Floor, Cambridge, Massachusetts 02142, USA~…}
}

@inproceedings{deng2024investigating,
 title={Investigating data contamination in modern benchmarks for large language models},
 author={Deng, Chunyuan and Zhao, Yilun and Tang, Xiangru and Gerstein, Mark and Cohan, Arman},
 booktitle={Proceedings of the 2024 Conference of the North American Chapter of the Association for Computational Linguistics: Human Language Technologies (Volume 1: Long Papers)},
 pages={8706--8719},
 year={2024}
}

@inproceedings{xiao2025restoring,
  title={Restoring calibration for aligned large language models: A calibration-aware fine-tuning approach},
  author={Xiao, Jiancong and Hou, Bojian and Wang, Zhanliang and Jin, Ruochen and Long, Qi and Su, Weijie J and Shen, Li},
  booktitle={International Conference on Machine Learning},
  year={2025}
}

@inproceedings{leng2025taming,
  title={Taming overconfidence in llms: Reward calibration in rlhf},
  author={Leng, Jixuan and Huang, Chengsong and Zhu, Banghua and Huang, Jiaxin},
  booktitle={International Conference on Learning Representations},
  volume={2025},
  year={2025}
}

@article{lee2025training,
  title={Training-free LLM Verification via Recycling Few-shot Examples},
  author={Lee, Dongseok and Hong, Jimyung and Kim, Dongyoung and Kim, Jaehyung},
  journal={arXiv preprint arXiv:2506.17251},
  year={2025}
}

@misc{makhija2025neuralbreadcrumbsmembershipinference,
      title={Neural Breadcrumbs: Membership Inference Attacks on LLMs Through Hidden State and Attention Pattern Analysis}, 
      author={Disha Makhija and Manoj Ghuhan Arivazhagan and Vinayshekhar Bannihatti Kumar and Rashmi Gangadharaiah},
      year={2025},
      eprint={2509.05449},
      archivePrefix={arXiv},
      primaryClass={cs.LG},
      url={https://arxiv.org/abs/2509.05449}, 
}

@misc{ranjan2026gdriftmiamembershipinference,
      title={G-Drift MIA: Membership Inference via Gradient-Induced Feature Drift in LLMs}, 
      author={Ravi Ranjan and Utkarsh Grover and Xiaomin Lin and Agoritsa Polyzou},
      year={2026},
      eprint={2604.00419},
      archivePrefix={arXiv},
      primaryClass={cs.LG},
      url={https://arxiv.org/abs/2604.00419}, 
}
\appendix
\newpage
\appendix
\twocolumn

\section{Details of Curated Datasets}\label{app:contam_data}
We provide examples of the curated evaluation and training dataset in Table~\ref{tab:contamination_setup} and Table~\ref{tab:dataset_examples}.
\setlength{\dbltextfloatsep}{4pt}

\begin{table*}[t]
\centering
\small
\renewcommand{\arraystretch}{1.15}

\begin{tabular*}{\textwidth}{@{\extracolsep{\fill}} l l c c l @{}}
\toprule

\multirow{2}{*}{\textbf{Split}}
& \multirow{2}{*}{\textbf{Source}}
& \multicolumn{2}{c}{\textbf{Composition}}
& \multirow{2}{*}{\textbf{Purpose}} \\

\cmidrule(lr){3-4}

&
& \textbf{Mem.}
& \textbf{Non-Mem.}
& \\

\midrule

\textbf{Evaluation}
& Open-source + AIME'26
& 30
& 30
& Evaluation \\

\textbf{Training}
& 30 members + 970 RL-MIA
& 30
& 970
& RL Analysis \\

\bottomrule
\end{tabular*}

\caption{
\textbf{Overview of the curated datasets.}
}
\label{tab:contamination_setup}

\end{table*}
\setlength{\dbltextfloatsep}{4pt}
\begin{table*}[t]
\centering
\small
\resizebox{\textwidth}{!}{
\renewcommand{\arraystretch}{0.9}

\begin{tabular}{p{2.3cm} p{10.2cm} p{1.1cm} p{0.9cm} p{2.2cm}}
\toprule
\textbf{data\_source} & \textbf{prompt} & \textbf{answer} & \textbf{member} & \textbf{metadata} \\
\midrule

\multicolumn{5}{l}{\textbf{Contamination evaluation samples}} \\
\midrule

\texttt{aime26} &
\texttt{[
\{role: system, content: When tackling complex reasoning tasks, you have access to actions such as ASSESS, ADVANCE, VERIFY, SIMPLIFY, SYNTHESIZE, PIVOT, and OUTPUT.\},
\{role: user, content: Patrick started walking at a constant rate from school to the park. Tanya ran 2 miles per hour faster than Patrick, Jose bicycled 7 miles per hour faster than Tanya, and all three arrived at the same time. Find m+n.\}
]} &
277 &
0 &
-- \\

\midrule

\texttt{numina\_amc\_aime} &
\texttt{[
\{role: system, content: When tackling complex reasoning tasks, you have access to actions such as ASSESS, ADVANCE, VERIFY, SIMPLIFY, SYNTHESIZE, PIVOT, and OUTPUT.\},
\{role: user, content: The highest price is \$8.50 and the lowest price is \$5.50. Calculate the percent by which the highest price is more than the lowest price.\}
]} &
70\% &
1 &
-- \\

\midrule

\multicolumn{5}{l}{\textbf{RL training sample}} \\
\midrule

\texttt{olympiads} &
\texttt{[
\{role: system, content: Your task is to follow a systematic, thorough reasoning process before providing the final solution. Structure your response into two sections: Thought and Solution. In the Thought section, present your reasoning using <think>\{thoughts\}</think>. In the Solution section, provide the final logical answer, optionally in \textbackslash boxed\{\} format.\},
\{role: user, content: Determine the largest even positive integer which cannot be expressed as the sum of two composite odd positive integers.\}
]} &
38 &
-- &
\texttt{\{style: "rule"\}} \\

\bottomrule
\end{tabular}
}

\caption{
\textbf{Samples from the curated contamination evaluation and RL training datasets.}
Each instance contains the data source, structured conversational prompt, ground-truth answer, membership label when applicable, and metadata annotations.
}
\label{tab:dataset_examples}
\end{table*}

\section{Implementation Details}

\subsection{Training Details}

We fine-tune the base model using Group Relative Policy Optimization (GRPO) within the VeRL~\citep{sheng2024hybridflow} training framework. Training is conducted for 2 epochs with a learning rate of $1\times10^{-6}$, train batch size 128, validation batch size 512, maximum prompt length 1024, and maximum response length 4096. We enable gradient checkpointing and dynamic batch sizing during optimization, with a per-GPU token budget of 16384 tokens. For rollout generation, we use vLLM~\citep{kwon2023efficient} with 4 sampled responses per prompt at temperature 1.0, while validation uses temperature 0.6. We do not apply explicit KL regularization during training. All training experiments are performed on $8\times$ NVIDIA A6000 GPUs.

\subsection{Inference Details}

We conduct two types of evaluation: reasoning evaluation and contamination evaluation.

\paragraph{Reasoning Evaluation.}\label{app:impl_reasoning}
 We conduct reasoning evaluation in Section~\ref{app:impl_reasoning} to test training robustness. Here, we generate $5$ sampled responses per example and report $\mathrm{pass}@5$, where a prediction is considered correct if at least one sampled response matches the ground-truth answer after extracting the final boxed or numeric answer. In addition, we compute the mean length-normalized token log-probability of the gold answer under the model:

\[
\frac{1}{T}\sum_{t=1}^{T}\log p(a_t \mid \mathrm{prompt}, a_{<t}),
\]

where $T$ denotes the answer length. Log-probabilities are computed using either a standard Transformers forward pass or vLLM-based prompt log-probability scoring.

\paragraph{Contamination Evaluation.}
Contamination evaluation follows a two-stage pipeline. First, we generate model responses together with token-level statistics such as log-probabilities and entropies. Second, we compute contamination detection scores using both output-based and representation-based methods.

The evaluated baselines include Min-K\%, PPL, CDD, Recall, and Self-Critique. For representation-based methods, LaRA constructs semantically related and perturbed variants of each question using gpt-4o-mini-generated (via OpenRouter API) ~\citep{openai2024gpt4omini, openrouter} paraphrases and incomplete-question variants, from which representation-level signals such as RSI/RSM and directional collapse are derived.
We evaluate member versus non-member separability using AUROC, TPR at fixed FPR (0.05). All inference and evaluation experiments are performed on $4\times$ NVIDIA RTX 3090 GPUs.
\begin{table*}[t]
\centering
\small
\setlength{\tabcolsep}{6pt}
\renewcommand{\arraystretch}{1.15}

\begin{tabular}{l|c|cc|cc}
\toprule
\multirow{2}{*}{\textbf{Model}} &
\multirow{2}{*}{\textbf{Overall Pass@5}} &
\multicolumn{2}{c|}{\textbf{Pass@5 (\%)}} &
\multicolumn{2}{c}{\textbf{LogP(answer $\mid$ prompt)}} \\
& & \textbf{Member} & \textbf{Non-member} & \textbf{Member} & \textbf{Non-member} \\
\midrule

\rowcolor{lightblue}
\textbf{Qwen2.5-Math-7B (Base)} 
& 15.0 (9/60)
& 26.7 (8/30)
& 3.3 (1/30)
& -2.08
& -2.44
\\
\midrule

\textbf{Eurus-2-7B-PRIME (Initial)}
& 18.3 (11/60)
& 33.3 (10/30)
& 3.3 (1/30)
& -8.54
& -7.88
\\

\textbf{Eurus-2-7B-PRIME (Epoch 1)}
& 18.3 (11/60)
& 36.7 (11/30)
& 0.0 (0/30)
& -9.06
& -8.50
\\

\textbf{Eurus-2-7B-PRIME (Epoch 2)}
& 18.3 (11/60)
& 36.7 (11/30)
& 0.0 (0/30)
& -8.85
& -9.12
\\

\midrule

\rowcolor{lightblue}
\textbf{Qwen2.5-Math-7B (Base)} 
& 10.0 (6/60)
& 20.0 (6/30)
& 0.0 (0/30)
& -2.07
& -2.44
\\
\midrule

\textbf{LIMR (Initial)}
& 8.0 (5/60)
& 17.0 (5/30)
& 0.0 (0/30)
& -2.48
& -2.92
\\

\textbf{LIMR (Epoch 1)}
& 13.3 (8/60)
& 23.3 (7/30)
& 3.0 (1/30)
& -2.44
& -2.90
\\

\textbf{LIMR (Epoch 2)}
& 10.0 (6/60)
& 20.0 (6/30)
& 0.0 (0/30)
& -2.43
& -2.87
\\






\bottomrule
\end{tabular}

\caption{
\textbf{Comparison between the RL-trained open-source model and its base model on member and non-member samples.}
}
\label{tab:member_nonmember_eval}

\end{table*}

\Needspace{10\baselineskip}
\section{Main Table with Confidence Intervals}
\label{app:main_table_confidence}

To quantify the statistical uncertainty arising from the limited evaluation
size, we report bootstrap confidence intervals for both AUC and
TPR@5\% FPR. Due to computational constraints, we conduct this analysis
using EURUS as a representative testbed. Each evaluation set contains
60 examples, consisting of 30 member and 30 non-member examples. We
perform 10,000 stratified bootstrap resamples with replacement, preserving
the number of member and non-member examples in each resample, using
random seed 42. For each resample, we recompute the detection threshold
and evaluate both metrics. We report 95\% percentile bootstrap confidence
intervals, while the point estimate corresponds to the metric computed on
the full evaluation set.

\begin{table}[t]
    \centering
    \scriptsize
    \setlength{\tabcolsep}{3pt}
    \renewcommand{\arraystretch}{1.08}
    \resizebox{\columnwidth}{!}{%
        \begin{tabular}{llcc}
            \toprule
            \textbf{Checkpoint}
            & \textbf{Method}
            & \shortstack{\textbf{AUC}\\\textbf{(95\% CI)}}
            & \shortstack{\textbf{TPR@5\% FPR}\\\textbf{(95\% CI)}} \\
            \midrule

            Initial
            & S\_LaRA
            & 0.628 [0.461, 0.784]
            & 0.192 [0.033, 0.444] \\

            Initial
            & Self-Critique + S\_LaRA
            & 0.682 [0.521, 0.829]
            & 0.154 [0.000, 0.654] \\
            \midrule

            Epoch 1
            & Self-Critique
            & 0.586 [0.436, 0.730]
            & 0.233 [0.054, 0.417] \\

            Epoch 1
            & S\_LaRA
            & 0.679 [0.519, 0.825]
            & 0.192 [0.040, 0.409] \\

            Epoch 1
            & Self-Critique + S\_LaRA
            & 0.671 [0.512, 0.813]
            & 0.269 [0.077, 0.519] \\
            \midrule

            Epoch 2
            & Self-Critique
            & 0.820 [0.706, 0.915]
            & 0.367 [0.179, 0.645] \\

            Epoch 2
            & S\_LaRA
            & 0.704 [0.546, 0.849]
            & 0.154 [0.000, 0.591] \\

            Epoch 2
            & Self-Critique + S\_LaRA
            & 0.858 [0.732, 0.961]
            & 0.231 [0.000, 0.920] \\
            \bottomrule
        \end{tabular}%
    }
    \caption{
        \textbf{Bootstrap point estimates and 95\% percentile confidence
        intervals on EURUS}, computed using 10,000 stratified bootstrap
        resamples. Confidence intervals are generally wider for
        TPR@5\% FPR because its threshold is determined by only a small
        number of non-member examples.
    }
    \label{tab:bootstrap_ci}
\end{table}

Overall, the bootstrap analysis confirms that statistical uncertainty is
non-negligible at the current evaluation scale, particularly for
TPR@5\% FPR. Nevertheless, the principal AUC-based observations remain
unchanged. In particular, at the Epoch~2 checkpoint, Self-Critique
combined with S\_LaRA achieves the highest AUC, while S\_LaRA remains
competitive across checkpoints. Given the instability of low-FPR
estimates at this sample size, however, small differences in
TPR@5\% FPR should not be interpreted as statistically meaningful.

\section{Evaluation on Same-Source Member and Non-Member Sets}
\label{app:same_source}

A potential confounding factor in contamination detection is a distributional difference between the member and non-member sets. 
In our primary evaluation, we mitigate this concern by restricting member examples to Olympiad-level mathematics problems comparable to AIME 2026, rather than sampling indiscriminately from the model's entire training corpus. 
Nevertheless, the two sets may still differ in their collection sources.

To more directly control for this factor, we conduct an additional evaluation following the same-source protocol of \citet{tao2025detecting}.
Specifically, we use the KK and SAT benchmarks from ~\citet{tao2025detecting}, for which both member and non-member examples are drawn from the same underlying dataset and therefore share the same collection procedure, problem domain, and overall distribution. 
Within each benchmark, the only intended distinction is whether an example was included in the model's RL training set. 
We evaluate checkpoints after the first and second RL training epochs for both EURUS and LIMR.

Tables~\ref{tab:same_source_eurus} and~\ref{tab:same_source_limr} report ROC AUC and TPR at 5\% FPR. 
For EURUS, \textsc{LaRA} and its combination with Self-Critique consistently achieve above-random ROC AUC across both benchmarks and training checkpoints.
On KK, the combined method obtains ROC AUCs of 0.739 and 0.699 after the first and second epochs, respectively. 
On SAT, \textsc{LaRA} achieves ROC AUCs of 0.622 and 0.689, while the combined method obtains 0.627 and 0.670. 
These results also indicate that the relative effectiveness of the methods can change over the course of RL training.

The results are particularly clear for LIMR. 
Across all four benchmark--checkpoint combinations, \textsc{LaRA} achieves ROC AUCs between 0.781 and 0.821, whereas Self-Critique remains close to random performance. 
Combining the two signals does not consistently improve upon \textsc{LaRA} alone for this model, suggesting that the output-level Self-Critique signal contributes limited complementary information in this setting.

Overall, \textsc{LaRA} continues to distinguish member from non-member examples when dataset source, collection procedure, and problem domain are controlled. 
These findings reduce the likelihood that its performance is driven primarily by dataset-specific artifacts or superficial domain shifts, and instead support the conclusion that its representation-level signals capture whether examples were encountered during RL training.

\begin{table}[t]
    \centering
    \small
    \setlength{\tabcolsep}{3pt}
    \resizebox{\columnwidth}{!}{%
    \begin{tabular}{llcc}
        \toprule
        \textbf{Run} & \textbf{Method}
        & \textbf{ROC AUC}
        & \textbf{TPR@5\% FPR} \\
        \midrule
        KK epoch 1
        & Self-Critique
        & 0.664 & 0.233 \\
        & \textsc{LaRA}
        & 0.641 & 0.192 \\
        & Self-Critique + \textsc{LaRA}
        & \textbf{0.739} & \textbf{0.385} \\
        \midrule
        KK epoch 2
        & Self-Critique
        & 0.619 & 0.200 \\
        & \textsc{LaRA}
        & \textbf{0.724} & \textbf{0.269} \\
        & Self-Critique + \textsc{LaRA}
        & 0.699 & \textbf{0.269} \\
        \midrule
        SAT epoch 1
        & Self-Critique
        & 0.579 & 0.167 \\
        & \textsc{LaRA}
        & 0.622 & 0.154 \\
        & Self-Critique + \textsc{LaRA}
        & \textbf{0.627} & \textbf{0.231} \\
        \midrule
        SAT epoch 2
        & Self-Critique
        & 0.604 & 0.100 \\
        & \textsc{LaRA}
        & \textbf{0.689} & 0.154 \\
        & Self-Critique + \textsc{LaRA}
        & 0.670 & \textbf{0.231} \\
        \bottomrule
    \end{tabular}%
    }
    \caption{\textbf{Same-source contamination detection results for EURUS.}
    Member and non-member examples are drawn from the same underlying
    benchmark. The best result within each run is shown in bold.}
    \label{tab:same_source_eurus}
\end{table}

\begin{table}[t]
    \centering
    \small
    \setlength{\tabcolsep}{3pt}
    \resizebox{\columnwidth}{!}{%
    \begin{tabular}{llcc}
        \toprule
        \textbf{Run} & \textbf{Method}
        & \textbf{ROC AUC}
        & \textbf{TPR@5\% FPR} \\
        \midrule
        KK epoch 1
        & Self-Critique
        & 0.456 & 0.000 \\
        & \textsc{LaRA}
        & \textbf{0.798} & \textbf{0.435} \\
        & Self-Critique + \textsc{LaRA}
        & 0.567 & 0.044 \\
        \midrule
        KK epoch 2
        & Self-Critique
        & 0.524 & 0.067 \\
        & \textsc{LaRA}
        & \textbf{0.798} & \textbf{0.480} \\
        & Self-Critique + \textsc{LaRA}
        & 0.624 & 0.080 \\
        \midrule
        SAT epoch 1
        & Self-Critique
        & 0.528 & 0.033 \\
        & \textsc{LaRA}
        & \textbf{0.821} & \textbf{0.520} \\
        & Self-Critique + \textsc{LaRA}
        & 0.656 & 0.240 \\
        \midrule
        SAT epoch 2
        & Self-Critique
        & 0.476 & 0.000 \\
        & \textsc{LaRA}
        & \textbf{0.781} & \textbf{0.160} \\
        & Self-Critique + \textsc{LaRA}
        & 0.598 & 0.080 \\
        \bottomrule
    \end{tabular}%
    }
    \caption{\textbf{Same-source contamination detection results for LIMR.}
    Member and non-member examples are drawn from the same underlying
    benchmark. The best result within each run is shown in bold.}
    \label{tab:same_source_limr}
\end{table}

\section{Additional Results on Figure Geometry}\label{app:figure_geometry}

\begin{figure}
    \includegraphics[width=\columnwidth]{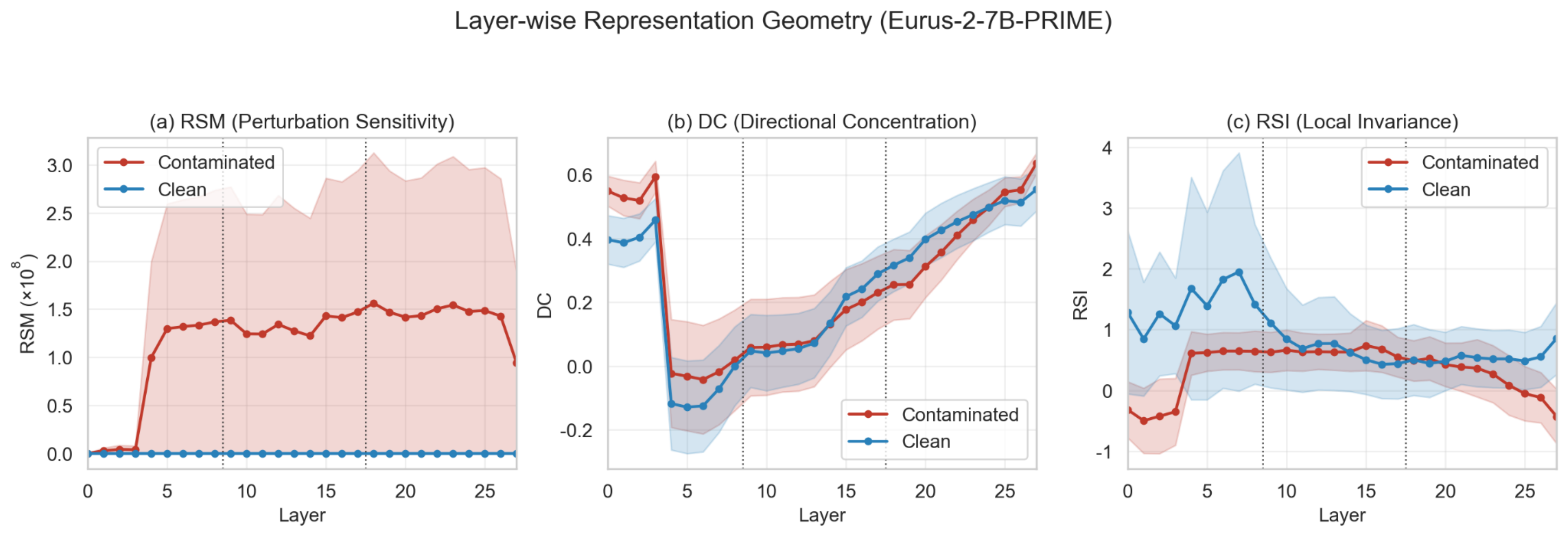}
    \includegraphics[width=\columnwidth]{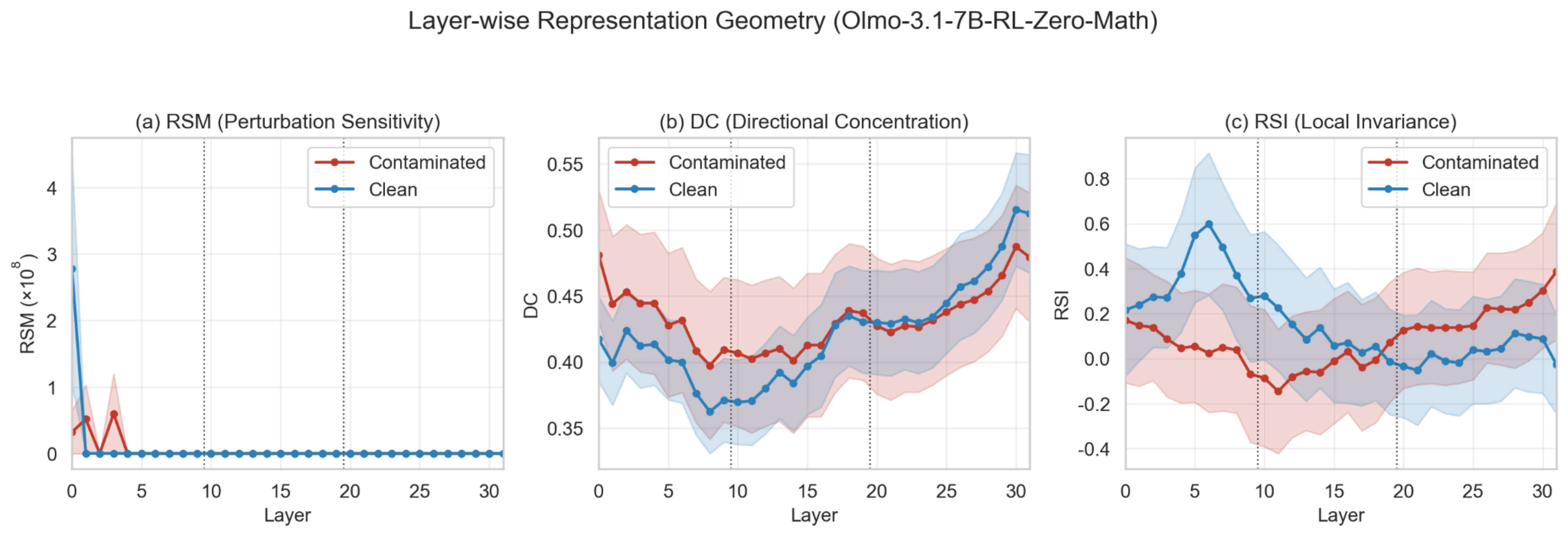}
    \caption{\textbf{Representation Geometry of Additional Models.}}
    \label{fig:repgeo_figure}
\end{figure}

Figure~\ref{fig:repgeo_figure} further shows that the layer-wise representation geometry patterns vary across RL-trained models while still exhibiting contamination-associated deviations. In \texttt{Eurus-2-7B-PRIME}, contaminated samples exhibit persistently elevated RSM values throughout depth with substantially larger magnitudes than clean samples, indicating strong perturbation sensitivity under information removal. Regarding DC, while the specific directional profiles vary across models and checkpoints, contaminated samples consistently exhibit deviations from the stable directional organization observed in clean samples, indicating disrupted and lower-dimensional perturbation dynamics. Exhibit comparatively lower RSI values than clean samples, suggesting induced local representation flexibility under paraphrastic perturbations.

In contrast, \texttt{Olmo-3.1-7B-RL-Zero-Math} exhibits weaker RSM separation, with both contaminated and clean samples remaining near zero across most layers. However, contamination-related deviations remain observable in DC and RSI, where contaminated samples maintain consistently higher directional concentration and smoother local variability patterns than clean samples. Overall, these results suggest that while the precise geometric behaviors vary across models, contamination consistently alters perturbation sensitivity, directional organization, and local representation variability.

\section{Further Analyses }
\subsection{Analyses of Layer-window Trends on Different Models}

\begin{figure}[t]
    \centering
    \includegraphics[width=\columnwidth]{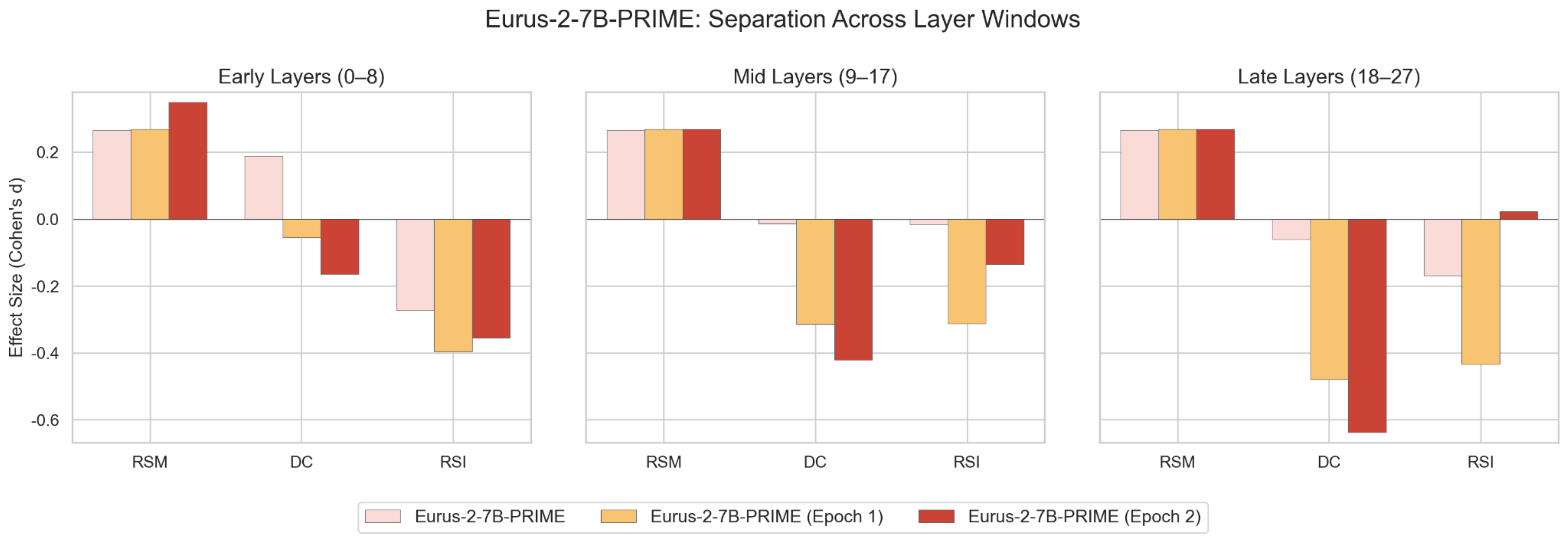}
    \captionsetup{width=\columnwidth, font=small}
    \caption{\textbf{Effect-size separation across early, mid, and late layer windows of
    \texttt{Eurus-2-7B-PRIME}.}
    Contaminated samples are consistently elevated on RSM but progressively lower
    on DC and RSI in mid-to-late layers, with the widened gap indicating that RL
    training amplifies a layer-selective representational signature of contamination.}
    \label{fig:figure_separation_combined}
\end{figure}

\paragraph{Analysis on \texttt{Eurus-2-7B-Prime}.}
Figure~\ref{fig:figure_separation_combined} shows contaminated-vs-clean separation (Cohen’s $d$) across early ($0$--$8$), mid ($9$--$17$), and late ($18$--$27$) layers over three RL checkpoints. Three consistent trends emerge. First, \textbf{RSM remains nearly unchanged across both depth and training}. All checkpoints show stable positive separation ($d\approx0.27$--$0.35$), indicating that perturbation sensitivity is largely preserved throughout RL fine-tuning. Second, \textbf{DC becomes increasingly negative as RL training progresses}, especially in deeper layers. While the initial checkpoint shows weak or near-zero separation, Epoch~1 and Epoch~2 exhibit progressively larger negative effects, with the strongest separation appearing in late layers ($d\approx-0.65$ at Epoch~2). This suggests that RL training amplifies directional-collapse behavior on contaminated samples, particularly in deeper representations. Third, \textbf{RSI is strongest in early and mid layers but weakens in late layers after training}. Early-layer separation remains consistently negative across checkpoints, whereas the late-layer signal gradually diminishes and nearly disappears by Epoch~2. This indicates that local-invariance differences are primarily concentrated in shallower representations. Overall, the figure reveals a clear layer-conditioned structure: RSM is stable across training, DC is progressively amplified by RL in deeper layers, and RSI is concentrated in earlier layers. These complementary trends motivate combining all three metrics in LaRA rather than relying on a single layer-wise signal.

\begin{figure}[t]
    \centering
    \includegraphics[width=\columnwidth]{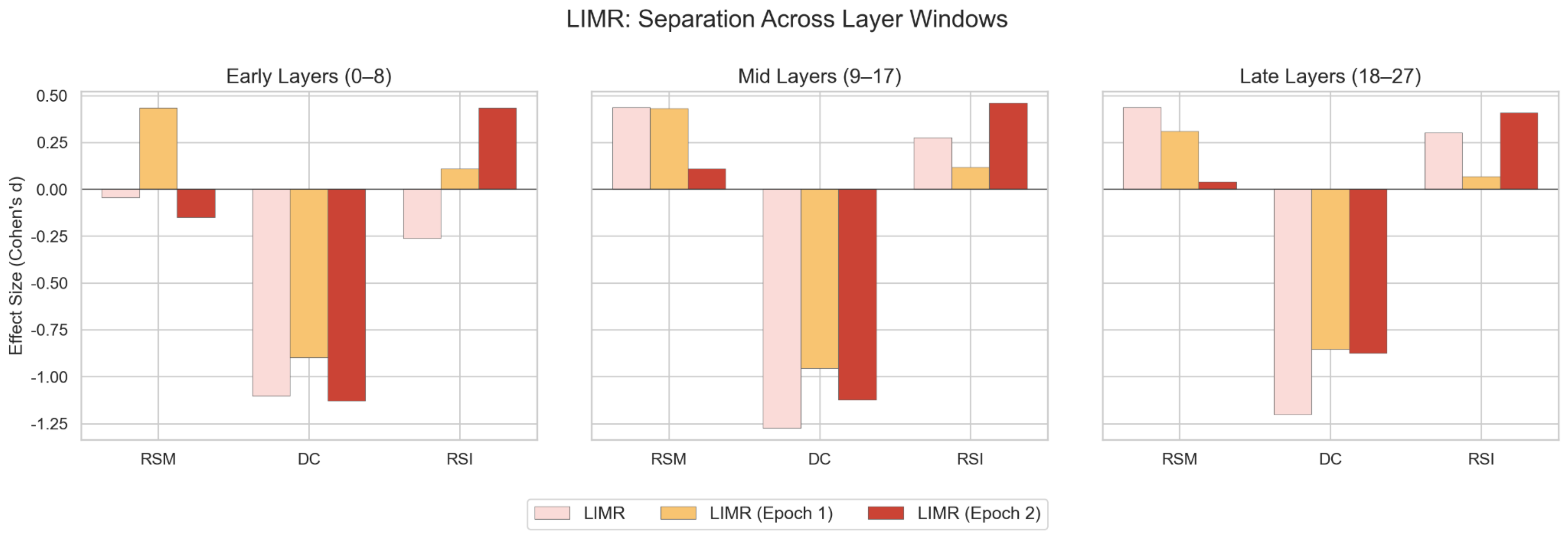}
    \captionsetup{width=\columnwidth, font=small}
    \caption{\textbf{Effect-size separation across early, mid, and late layer windows of
    \texttt{LIMR}.}
    Contaminated samples exhibit consistently lower DC across all layer windows,
    while RSI becomes increasingly positive in mid-to-late layers as RL training
    progresses. In contrast to Eurus-2-7B-PRIME, the separation dynamics in LIMR
    are dominated by strong directional concentration differences together with
    progressively amplified RSI gaps, indicating that RL training induces a
    distinct layer-dependent representation geometry signature of contamination
    in LIMR.}
    \label{fig:figure_separation_limr_combined}
\end{figure}

\paragraph{Analysis on \texttt{LIMR}.}
Figure~\ref{fig:figure_separation_limr_combined} shows contaminated-vs-clean separation (Cohen’s $d$) across early ($0$--$8$), mid ($9$--$17$), and late ($18$--$27$) layers over three RL checkpoints. Several distinct trends emerge. First, \textbf{RSM remains consistently positive across all layer groups}, indicating that contaminated samples exhibit larger representation shifts under perturbation throughout training. However, unlike the relatively stable behavior observed in \texttt{Eurus-2-7B-Prime}, the magnitude of separation decreases after RL training, particularly in deeper layers. Early and mid layers initially show strong separation ($d\approx0.43$), but this gradually weakens by Epoch~2, especially in the late layers where the effect becomes nearly negligible. This suggests that RL training partially suppresses perturbation-sensitive geometry in LIMR. Second, \textbf{DC exhibits the strongest and most stable separation across all checkpoints}. All layer groups consistently show large negative effects ($d\approx-0.85$ to $-1.25$), with mid and late layers displaying the strongest separation. Moreover, RL training slightly amplifies this negative separation in Epoch~1 and Epoch~2, indicating that contaminated representations become increasingly directionally collapsed during RL optimization. Compared to the other metrics, DC provides the clearest and most persistent distinction between contaminated and clean samples across depth. Third, \textbf{RSI shows progressively stronger positive separation as RL training proceeds}, particularly in mid and late layers. While the initial checkpoint exhibits only moderate separation, Epoch~2 produces substantially larger effects ($d\approx0.40$--$0.45$), especially beyond the middle layers. This indicates that RL training amplifies local-invariance differences between contaminated and clean samples, causing contaminated representations to become increasingly invariant under local perturbations. Overall, LIMR exhibits a complementary layer-wise structure distinct from \texttt{Eurus-2-7B-Prime}: RSM weakens during RL training, DC remains consistently dominant across all depths, and RSI becomes progressively amplified in deeper layers. These trends suggest that RL optimization in LIMR increasingly concentrates contaminated representations into geometrically collapsed yet locally invariant structures, motivating the joint use of RSM, DC, and RSI in LaRA for robust contamination detection.

\subsection{Aggregation over Different Layer-windows.}

\begin{figure}[t]
    \centering
    \includegraphics[width=\columnwidth]{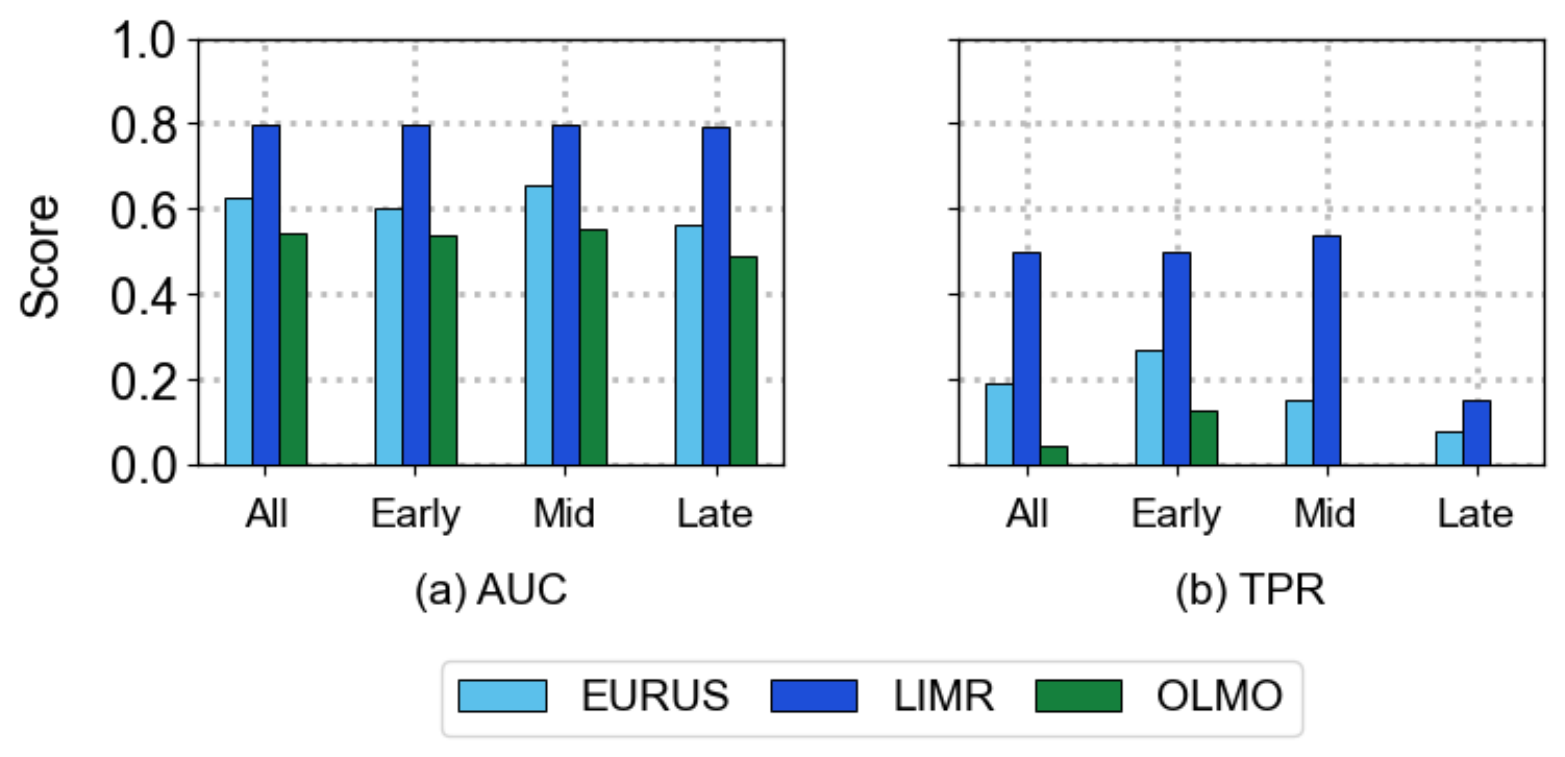}
    \captionsetup{width=\columnwidth, font=small}
    \caption{\textbf{Results over different layer windows on three models.}}
    \label{fig:window_bar}
\end{figure}

Figure~\ref{fig:window_bar} compares contamination detection performance when aggregating representation statistics over different layer windows (Early, Mid, and Late) as well as over all layers jointly. We report both AUC and TPR@FPR=5\% across three RL-trained models. A clear and consistent trend emerges: \textbf{the relative contamination-separation behavior remains largely stable across different layer regions}. 

Across all models, the performance ordering is highly consistent regardless of which layer window is used. LIMR consistently achieves the strongest separation across every window, obtaining nearly identical AUC values ($\approx0.8$) for Early, Mid, and Late layers, along with the highest TPR values throughout. EURUS exhibits moderate but stable performance across all windows, while OLMO consistently shows weaker separation. Importantly, no single layer region dominates the results; instead, contamination-related geometric signals persist throughout the network depth. The results demonstrate that contamination-related representation geometry is remarkably stable across layer depth. Rather than arising from isolated layers, the separation signal persists throughout the network, motivating the use of layer-window aggregation as a robust and model-agnostic strategy for contamination detection.

\subsection{Metric Ablations on Additional Model}
We also conduct an additional ablation study on another model, \texttt{Eurus-2-7B-Prime} in Table~\ref{tab:metric_ablation_eurus}. The results show that $S_{LaRA}$ is relatively robust to ablations, as shown in \texttt{LIMR} results in the main ablation studies.

\subsection{Analysis on the Evolution of Representation Geometry across RL Training Stages}
\begin{figure}[t]
    \centering
    \includegraphics[width=\columnwidth]{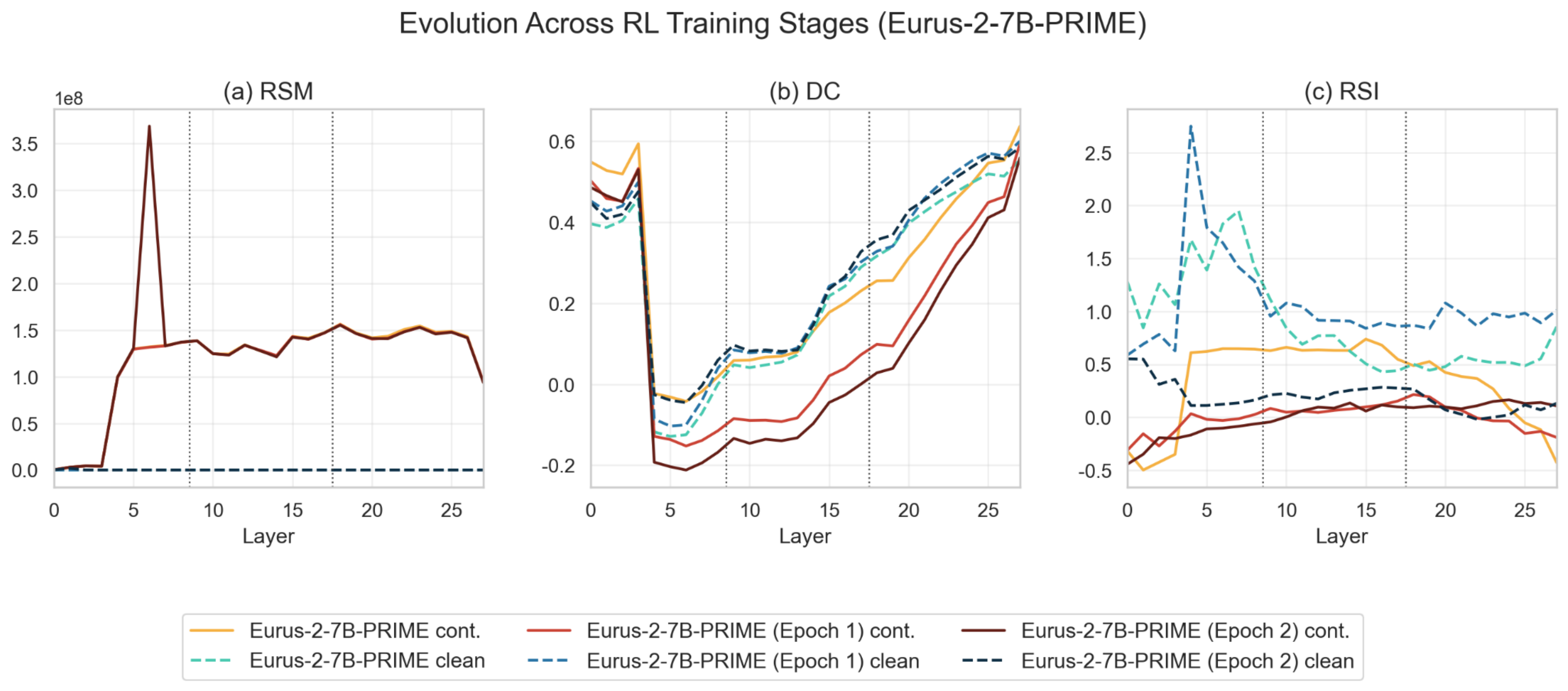}
    \captionsetup{width=\columnwidth, font=small}
    \caption{\textbf{Evolution of layer-wise representation geometry across RL training stages.}
    RL progressively amplifies geometric deviations in each metric
    (\textbf{RSM, DC, and RSI}) between clean and contaminated samples over epochs.}
    \label{fig:figure_evolution_training}
    \vspace{2.0em}
\end{figure}

Figure~\ref{fig:figure_evolution_training} illustrates how RL training progressively reshapes the layer-wise representation geometry of contaminated samples relative to clean samples in \texttt{Eurus-2-7B-PRIME}. Across all three metrics, the separation between clean and contaminated trajectories becomes increasingly pronounced as training advances from the initial checkpoint to Epochs 1 and 2. In particular, \textbf{RSM} exhibits the strongest divergence, where contaminated samples show dramatically amplified sensitivity beginning in early-middle layers and remaining consistently elevated throughout deeper layers, while clean samples stay nearly flat across all stages. \textbf{DC} further reveals that RL training induces increasingly distinct directional structure: contaminated samples transition from negative or near-zero values in middle layers to strongly positive values in deeper layers, whereas clean samples maintain smoother and more stable trajectories. Similarly, \textbf{RSI} demonstrates that contaminated samples undergo substantial geometric instability during RL training, especially in shallow layers where sharp spikes emerge after training, while clean samples retain comparatively moderate and stable behavior. Importantly, these trends remain consistent across layer depth and training stages, suggesting that RL fine-tuning systematically amplifies hidden-state geometric irregularities associated with memorized or contaminated data rather than producing isolated layer-specific effects.

\begin{table}[t]
    \centering
    \small
    \renewcommand{\arraystretch}{1.05}
    \setlength{\tabcolsep}{3.2pt}

    \resizebox{\columnwidth}{!}{%
        \begin{tabular}{ccc|cc cc cc}
            \toprule

            \textbf{RSM}
            & \textbf{DC}
            & \textbf{RSI}
            & \multicolumn{2}{c|}{\textbf{Init}}
            & \multicolumn{2}{c|}{\textbf{E1}}
            & \multicolumn{2}{c}{\textbf{E2}} \\

            \cmidrule(lr){4-5}
            \cmidrule(lr){6-7}
            \cmidrule(lr){8-9}

            & & &
            \textbf{AUC} & \textbf{TPR}
            & \textbf{AUC} & \textbf{TPR}
            & \textbf{AUC} & \textbf{TPR} \\

            \midrule

            {\color{red}\ding{55}}
            & {\color{red}\ding{55}}
            & {\color{green!60!black}\ding{51}}
            & 0.59 & 0.19
            & 0.49 & 0.00
            & 0.54 & 0.08 \\

            {\color{red}\ding{55}}
            & {\color{green!60!black}\ding{51}}
            & {\color{red}\ding{55}}
            & 0.68 & 0.12
            & 0.74 & 0.35
            & 0.62 & 0.04 \\

            {\color{green!60!black}\ding{51}}
            & {\color{red}\ding{55}}
            & {\color{red}\ding{55}}
            & 0.59 & 0.27
            & 0.53 & 0.23
            & 0.52 & 0.27 \\

            \midrule

            {\color{red}\ding{55}}
            & {\color{green!60!black}\ding{51}}
            & {\color{green!60!black}\ding{51}}
            & 0.68 & 0.19
            & 0.69 & 0.35
            & 0.61 & 0.08 \\

            {\color{green!60!black}\ding{51}}
            & {\color{red}\ding{55}}
            & {\color{green!60!black}\ding{51}}
            & 0.62 & 0.23
            & 0.54 & 0.23
            & 0.56 & 0.23 \\

            {\color{green!60!black}\ding{51}}
            & {\color{green!60!black}\ding{51}}
            & {\color{red}\ding{55}}
            & 0.67 & 0.23
            & 0.73 & 0.15
            & 0.62 & 0.19 \\

            \midrule

            \rowcolor{lightblue}
            {\color{green!60!black}\ding{51}}
            & {\color{green!60!black}\ding{51}}
            & {\color{green!60!black}\ding{51}}
            & 0.68 & 0.19
            & 0.70 & 0.15
            & 0.63 & 0.19 \\

            \bottomrule
        \end{tabular}%
    }

    \captionsetup{width=\columnwidth}
    \caption{\textbf{Ablations on Eurus across RL epochs.}}
    \label{tab:metric_ablation_eurus}
\end{table}\label{metric_ablation_eurus}

\section{Justifications of the Scaling Factor Metric in Contamination Detection Protocol}\label{app:standard-scaling-factor}

The factor $1.4826 = 1/\Phi^{-1}(0.75)$ is the standard
Fisher-consistency constant used in robust statistics to make the median
absolute deviation (MAD) a consistent estimator of the standard deviation
under a Gaussian reference~\citep{hampel1986robust, huber2009robust,
rousseeuw1993alternatives}. Specifically, if
$X \sim \mathcal{N}(\mu,\sigma^2)$, then
$\operatorname{median}|X-\mu| = \sigma\,\Phi^{-1}(0.75)$, so multiplying
the empirical MAD by
$1/\Phi^{-1}(0.75) \approx 1.4826$ recovers $\sigma$ in the
large-sample limit. Two properties of this scaling are particularly important for our
protocol. First, it places the robust scale estimate on the same
numerical units as the ordinary sample standard deviation, so the
standardized deviations $z_{m,\ell}(x)$, the metric-specific alignments
$\hat z_{m,\ell}(x)$, and the aggregated score
$S_{\mathrm{LaRA}}(x)$ retain their interpretation as approximate
Gaussian-style $z$-scores. Consequently, replacing the standard
deviation with a robust scale estimator does not implicitly retune
downstream thresholds or alter the semantic interpretation of the score.

Second, unlike the sample standard deviation, which has a $0\%$
breakdown point and can be driven arbitrarily large by a single extreme
outlier, the MAD achieves a $50\%$ breakdown point and therefore remains
stable even when the clean-reference pool
$\mathcal{D}^{\mathrm{clean}}$ contains a substantial fraction of
atypical samples~\citep{hampel1986robust, rousseeuw1987robust}. This is
particularly relevant for representation-geometry metrics, whose raw
distributions are often heavy-tailed even after signed-$\log(1+|\cdot|)$
compression. The resulting robustness-efficiency trade-off is well
established in the robust statistics literature: while the MAD is less
asymptotically efficient than the sample standard deviation under
perfectly Gaussian noise, it provides substantially improved stability
under contamination and heavy-tailed deviations
~\citep{huber2009robust, maronna2019robust}.

\section{Metrics and Baselines}\label{app:metrics_and_baselines}
\subsection{Metrics}
\textbf{ROC-AUC} measures the model’s ability to distinguish between member and non-member samples across all possible decision thresholds. It captures the overall separability of the two classes and is threshold-independent, making it a robust indicator of detection quality. \textbf{TPR@FPR=5\%} reports the true positive rate (i.e., correctly identified members) when the false positive rate (i.e., non-members incorrectly flagged as members) is fixed at 5\%. This reflects performance in the low false-positive regime, which is critical in contamination detection where incorrectly labeling clean samples as contaminated is costly. These metrics provide a comprehensive evaluation of global discrimination ability (ROC-AUC) and practical operating performance under strict error constraints (TPR@FPR=5\%). \

\subsection{Baselines}
We use six baselines to compare against our contamination detection protocol. (1) Recall~\citep{xie2024recall}, which probes memorization by measuring the model’s ability to regenerate ground-truth answers under controlled prompting; (2) CDD~\citep{dong2024generalization}, which detects contamination via discrepancies in model predictions under input or prompt perturbations, based on the intuition that memorized samples are less sensitive to such changes; (3) Min-K\% Prob~\citep{shi2023detecting}, a likelihood-based metric that averages the log-probability over the lowest-probability tokens in a sequence, assuming memorized samples exhibit fewer low-confidence tokens; (4) Min-K\%++~\citep{zhang2024min}, which extends Min-K\% with improved normalization and calibration for greater robustness across settings; (5) PPL~\citep{gonen2023demystifying}, which measures sequence-level likelihood via perplexity, where unusually low values indicate potential memorization; and (6) Self-Critique~\citep{tao2025detecting}, which leverages the model’s own reflective reasoning to assess contamination based on the confidence and consistency of its self-evaluation.

\section{Algorithm}
Details of the LaRA algorithm is in Appendix~\ref{alg:algorithm}.
\begin{algorithm*}[t]
\small

\caption{LaRA: Per-sample Layer-wise Representation Geometry Extraction}
\label{alg:repstiff}

\KwIn{
Original question $q_0$;
similar-question generator $\textsc{SimilarGen}$;
importance-based blanking operator $\textsc{BlankImportant}$ producing $k$ \texttt{[BLANK]} tokens;
LLM paraphrase generator $\textsc{VariantGen}$ that preserves \texttt{[BLANK]} positions;
mean-pooled hidden-state extractor $h_\ell(\cdot)$;
layer set $\mathcal{L}=\{0,1,\ldots,L-1\}$ (every transformer layer);
number of similar questions $K$;
number of paraphrase variants $M$;
number of \texttt{[BLANK]} tokens $k$;
numerical floor $\epsilon$
}

\KwOut{
Per-layer geometric scores
$\{\textnormal{RSM}_\ell,\textnormal{DC}_\ell,\textnormal{RSI}_\ell\}_{\ell\in\mathcal{L}}$
}

\vspace{0.3em}

$\{q_1,\dots,q_K\} \leftarrow \textsc{SimilarGen}(q_0)$\;

$\mathcal{Q} \leftarrow \{q_0,q_1,\dots,q_K\}$\;

\vspace{0.3em}

\ForEach{$q_i \in \mathcal{Q}$}{

    $q_i^{-} \leftarrow \textsc{BlankImportant}(q_i,\,k)$\;

    $\{v_{i,1},\dots,v_{i,M}\}
    \leftarrow \textsc{VariantGen}(q_i^{-})$\;
}

\vspace{0.3em}

\ForEach{$\ell \in \mathcal{L}$}{

    \tcc{(1) Representation Shift Magnitude $\rightarrow$ RSM}

    \For{$i=0$ \KwTo $K$}{

        $u_i \leftarrow h_\ell(q_i)$\;

        $w_i \leftarrow h_\ell(q_i^{-})$\;

        $\Delta_i \leftarrow u_i - w_i$\;

        $S_i \leftarrow \|\Delta_i\|_2$\;
    }

    $\mu_S \leftarrow
    \frac{1}{K}\sum_{i=1}^{K} S_i$\tcp*{similars only}

    $\sigma_S \leftarrow
    \sqrt{
    \frac{1}{K-1}
    \sum_{i=1}^{K}(S_i-\mu_S)^2
    }$\tcp*{sample std}

    $\textnormal{RSM}_\ell \leftarrow
    \dfrac{S_0-\mu_S}{\sigma_S+\epsilon}$\;

    \vspace{0.4em}

    \tcc{(2) Directional Collapse $\rightarrow$ DC}

    $\bar{s}_\ell \leftarrow
    \frac{1}{K}\sum_{i=1}^{K}\Delta_i$\tcp*{mean shift over similars}

    $\textnormal{DC}_\ell \leftarrow
    \dfrac{
    \Delta_0^\top \bar{s}_\ell
    }{
    (\|\Delta_0\|_2+\epsilon)
    (\|\bar{s}_\ell\|_2+\epsilon)
    }$\tcp*{cosine similarity}

    \vspace{0.4em}

    \tcc{(3) Representation Stability Index $\rightarrow$ RSI}

    \For{$i=0$ \KwTo $K$}{

        \For{$m=1$ \KwTo $M$}{

            $\phi_{i,m} \leftarrow h_\ell(v_{i,m})$\;
        }

        $\bar{\phi}_i \leftarrow
        \frac{1}{M}
        \sum_{m=1}^{M}\phi_{i,m}$\;

        $R_i \leftarrow
        \frac{1}{M}
        \sum_{m=1}^{M}
        \|\phi_{i,m}-\bar{\phi}_i\|_2$\tcp*{mean L2 distance}
    }

    $\mu_R \leftarrow
    \frac{1}{K}\sum_{i=1}^{K} R_i$\;

    $\sigma_R \leftarrow
    \sqrt{
    \frac{1}{K-1}
    \sum_{i=1}^{K}(R_i-\mu_R)^2
    }$\;

    $\textnormal{RSI}_\ell \leftarrow
    \dfrac{R_0-\mu_R}{\sigma_R+\epsilon}$\;
}

\vspace{0.3em}

\Return{
$\{
\textnormal{RSM}_\ell,
\textnormal{DC}_\ell,
\textnormal{RSI}_\ell
\}_{\ell\in\mathcal{L}}$
}\;

\end{algorithm*}\label{alg:algorithm}
\section{Details of Generating Similar and Perturbed Questions}\label{app:similar_question_prompt}

\subsection{Prompts}

To analyze representation dynamics under controlled perturbations, we use a three-stage prompt pipeline consisting of: (1) generating structurally similar questions, (2) identifying removable key information, and (3) generating paraphrased perturbation variants. 

As shown in Table ~\ref{tab:similar_prompt}, we first generate semantically similar math problems that preserve the same reasoning structure and difficulty while modifying numerical values. This produces structurally matched control groups for representation comparison. Table ~\ref{tab:perturbed_prompt} and ~\ref{tab:variant_prompt_initial} illustrates the prompt used to generate paraphrased variants of perturbed questions while preserving the exact position and semantic role of the \texttt{[BLANK]} placeholder. These variants enable measurement of local representation variability for computing RSI. Other prompts used to make the variants are in Table~\ref{tab:variant_prompt_1}, ~\ref{tab:variant_prompt_2}, and ~\ref{tab:variant_prompt_3}.

\subsection{Examples of the Generated Questions}
We present the examples of the original and generated questions to qualitatively show their generations in Table~\ref{tab:combined_questions}. Other questions generated by other perturbation techniques are in Table~\ref{tab:combined_questions_1}, ~\ref{tab:combined_questions_2}, and ~\ref{tab:combined_questions_3}.

\section{Usage of AI Assistants}
In preparing this work, we used AI-based writing assistants to improve sentence structure, correct
grammatical errors, and enhance overall readability. These tools were employed solely for language
refinement and did not contribute to the development of technical content, research methodology, or experimental analysis. All scientific ideas, results, and conclusions presented in the paper were conceived and authored entirely by the researchers. Use of AI assistance was restricted to editorial purposes and did not affect the originality or intellectual contributions of the work.

\FloatBarrier
\begin{table*}[t]
\centering
\small
\renewcommand{\arraystretch}{1.15}

\begin{tabular}{p{3cm} p{12cm}}
\toprule
\textbf{Settings} & \textbf{Content} \\
\midrule

Similar Questions &
\begin{ttfamily}
You are a math problem generator. Given an original math problem, create \{num\_questions\} similar problems that:

1. Follow the EXACT same structure and solution method as the original
2. Use DIFFERENT numerical values (change all numbers to make the problem unique)
3. Maintain the same difficulty level
4. Have the same type of solution approach
5. Are valid, solvable problems

Original Problem:

\{original\_question\}

Generate \{num\_questions\} similar problems. For each problem:
- Change ALL numerical values to create unique scenarios
- Keep the problem structure and mathematical concepts identical
- Ensure the problem remains solvable and realistic
- Make sure the new numbers create valid mathematical relationships

Output ONLY a JSON array of \{num\_questions\} similar problems, where each element is a string containing the full problem text. Do not include solutions or explanations, only the problems.

Format your response as:

["{}"Problem 1 text here...", "{}"Problem 2 text here..."]
\end{ttfamily}
\\

\bottomrule
\end{tabular}

\caption{Prompt used for generating similar math questions.}
\label{tab:similar_prompt}
\end{table*}

\begin{table*}[t]
\centering
\small
\renewcommand{\arraystretch}{1.15}

\begin{tabular}{p{3cm} p{12cm}}
\toprule
\textbf{Settings} & \textbf{Content} \\
\midrule

Perturbed Questions &
\begin{ttfamily}
You are a question editor that identifies key information to remove from math problems.

Given a math problem, identify ONE key piece of information that should be removed. Describe this information in a way that can be consistently applied to similar problems.

For example:
- "the total number of residents/people"
- "the initial quantity"
- "the final result value"
- "the time duration"
- "the distance measurement"

Original Problem:

\{original\_question\}

Output ONLY a short description of what information type should be removed (e.g., "the total number of residents"). Do not include the actual value or explain why, just describe the information type in 5-10 words.
\end{ttfamily}
\\

\bottomrule
\end{tabular}

\caption{Prompt used for identifying removable information in math questions.}
\label{tab:perturbed_prompt}
\end{table*}

\begin{table*}[t]
\centering
\small
\renewcommand{\arraystretch}{1.15}

\begin{tabular}{p{3cm} p{12cm}}
\toprule
\textbf{Settings} & \textbf{Content} \\
\midrule

Perturbed Variants &
\begin{ttfamily}
You are a text rewriter that creates paraphrased versions of math problems.

Given an incomplete math problem with [BLANK] placeholders, create \{num\_variants\} paraphrased versions that:
1. Preserve the EXACT position and meaning of [BLANK] - do NOT move or change [BLANK]
2. Use different wording and phrasing while maintaining the same mathematical meaning
3. Keep the same structure and logical flow
4. Do NOT reveal what the blank should be
5. Maintain all mathematical relationships and constraints

Incomplete Problem:

\{incomplete\_question\}

Output ONLY a JSON array of \{num\_variants\} strings with the paraphrased versions. Do not include explanations or notes.
\end{ttfamily}
\\

\bottomrule
\end{tabular}

\caption{Prompt used for generating perturbed paraphrased variants.}
\label{tab:variant_prompt_initial}
\end{table*}

\begin{table*}[t]
\centering
\small
\renewcommand{\arraystretch}{1.15}

\begin{tabular}{p{3cm} p{12cm}}
\toprule
\textbf{Setting} & \textbf{Content} \\
\midrule

Perturbation Analysis \\ (Variable Renaming) &
\begin{minipage}[t]{\linewidth}
\ttfamily
You edit math problems for robustness testing.

For each problem, replace EXACTLY ONE single-letter variable (e.g. x, y, n, t) with a random common English noun
(not a single letter). Example: replace "x" with "widget" everywhere that variable appears in the problem, or only
once if the variable appears once --- pick one occurrence if multiple variables exist.

- Do NOT use [BLANK].
- Keep all numbers and the problem structure unchanged except for that substitution.

Problems: \{numbered\}

Output ONLY a JSON array of strings with the same length and order as the input.
\end{minipage}
\\

\bottomrule
\end{tabular}

\caption{Prompt used for perturbation analysis through renaming variable.}
\label{tab:variant_prompt_1}

\end{table*}

\begin{table*}[t]
\centering
\small
\renewcommand{\arraystretch}{1.15}

\begin{tabular}{p{3cm} p{12cm}}
\toprule
\textbf{Setting} & \textbf{Content} \\
\midrule

Perturbation Analysis \\ (Number Replacement) &
\begin{minipage}[t]{\linewidth}
\ttfamily
You edit math problems for robustness testing.

For each problem, change EXACTLY ONE numeric literal to a different numeric literal.
- Do NOT use [BLANK].
- Do NOT change anything else (wording, variables, structure).
- The new number must differ from the original.

Problems:
{numbered}

Output ONLY a JSON array of strings with the same length and order as the input.
\end{minipage}
\\

\bottomrule
\end{tabular}

\caption{Prompt used for perturbation analysis through number replacement.}
\label{tab:variant_prompt_2}

\end{table*}

\begin{table*}[t]
\centering
\small
\renewcommand{\arraystretch}{1.15}

\begin{tabular}{p{3cm} p{12cm}}
\toprule
\textbf{Setting} & \textbf{Content} \\
\midrule

Perturbation Analysis \\ (Insert Distractor) &
\begin{minipage}[t]{\linewidth}
\ttfamily
You edit math problems for robustness testing.

For each problem, insert EXACTLY ONE short distractor sentence (one clause) that is irrelevant or misleading but
plausible-sounding. Place it naturally in the problem text.
- Do NOT use [BLANK].
- Do not change the original numbers or the core question being asked.

Problems:
{numbered}

Output ONLY a JSON array of strings with the same length and order as the input.
\end{minipage}
\\

\bottomrule
\end{tabular}

\caption{Prompt used for perturbation analysis through inserting distractor.}
\label{tab:variant_prompt_3}

\end{table*}
\FloatBarrier

\begin{table*}[t]
\centering
\small
\renewcommand{\arraystretch}{1.2}

\begin{tabular}{p{3cm} p{12cm}}
\toprule
\textbf{Settings} & \textbf{Content} \\
\midrule

Original Question &
\begin{ttfamily}
A plane contains points $A$ and $B$ with $AB = 1$. Point $A$ is rotated in the plane counterclockwise through an acute angle $\theta$ around point $B$ to point $A^\prime$. Then $B$ is rotated in the plane clockwise through angle $\theta$ around point $A^\prime$ to point $B^\prime$. Suppose that $AB^\prime = \frac{4}{3}$. The value of $\cos \theta$ can be written as $\frac{m}{n}$, where $m$ and $n$ are relatively prime positive integers. Find $m+n$.

Present the answer in LaTeX format: $\boxed{\text{Your answer}}$
\end{ttfamily}
\\
\midrule

Similar Question &
\begin{ttfamily}
A plane contains points $A$ and $B$ with $AB = \color{red}2$. Point $A$ is rotated in the plane counterclockwise through an acute angle $\theta$ around point $B$ to point $A^\prime$. Then $B$ is rotated in the plane clockwise through angle $\theta$ around point $A^\prime$ to point $B^\prime$. Suppose that $AB^\prime = \frac{5}{4}$. The value of $\cos \theta$ can be written as $\frac{m}{n}$, where $m$ and $n$ are relatively prime positive integers. Find $m+n$.
\end{ttfamily}
\\
\midrule

Perturbed Question &
\begin{ttfamily}
A plane contains points $A$ and $B$ with $AB = \color{red}[BLANK]$. Point $A$ is rotated in the plane counterclockwise through an acute angle $\theta$ around point $B$ to point $A^\prime$. Then $B$ is rotated in the plane clockwise through angle $\theta$ around point $A^\prime$ to point $B^\prime$. Suppose that $AB^\prime = \frac{4}{3}$. The value of $\cos \theta$ can be written as $\frac{m}{n}$, where $m$ and $n$ are relatively prime positive integers. Find $m+n$.
\end{ttfamily}
\\
\midrule

Perturbed Variant &
\begin{ttfamily}
In a plane, there are points $A$ and $B$ such that $AB = [BLANK]$. Point $A$ is rotated counterclockwise around point $B$ through an acute angle $\theta$, resulting in point $A^\prime$. Next, point $B$ is rotated clockwise around point $A^\prime$ through the same angle $\theta$, leading to point $B^\prime$. It is given that $AB^\prime = \frac{9}{8}$. The value of $\cos \theta$ can be expressed as $\frac{m}{n}$, where $m$ and $n$ are positive integers with no common factors. Determine the sum $m+n$.
\end{ttfamily}
\\

\bottomrule
\end{tabular}

\caption{Examples of Target, Similar, and Perturbed Questions}
\label{tab:combined_questions}
\end{table*}
\FloatBarrier
\begin{table*}[t]
\centering
\small
\renewcommand{\arraystretch}{1.2}

\begin{tabular}{p{3cm} p{12cm}}
\toprule
\textbf{Settings} & \textbf{Content} \\
\midrule

Original Question &
\begin{minipage}[t]{12cm}\ttfamily
A plane contains points $A$ and $B$ with $AB = 1$. Point $A$ is rotated in the plane counterclockwise through an acute angle $\theta$ around point $B$ to point $A^{/\\prime}$. Then $B$ is rotated in the plane clockwise through angle $\theta$ around point $A^{/\\prime}$ to point $B^{/\\prime}$. Suppose that $AB^{/\\prime} = \frac{4}{3}$. The value of $\cos \theta$ can be written as $\frac{m}{n}$, where $m$ and $n$ are relatively prime positive integers. Find $m+n$.

Present the answer in LaTeX format: $\boxed{\text{Your answer}}$
\end{minipage}
\\
\midrule

Similar Question &
\begin{minipage}[t]{12cm}\ttfamily
A plane contains points $A$ and $B$ with $AB = \color{red}2$. Point $A$ is rotated in the plane counterclockwise through an acute angle $\\theta$ around point $B$ to point $A'$. Then $B$ is rotated in the plane clockwise through angle $\\theta$ around point $A'$ to point $B'$. Suppose that $AB' = 3$. The value of $\\cos \\theta$ can be written as $\\frac{m}{n}$, where $m$ and $n$ are relatively prime positive integers. Find $m + n$.

Present the answer in LaTeX format: $\boxed{\text{Your answer}}$
\end{minipage}
\\
\midrule

Perturbed Question &
\begin{minipage}[t]{12cm}\ttfamily
A plane contains points $A$ and $B$ with $AB = 1$. Point $A$ is rotated in the plane counterclockwise through an acute angle $\theta$ around point $B$ to point $\color{red}A'$. Then $B$ is rotated in the plane clockwise through angle $\theta$ around point $\color{red}A'$ to point $\color{red}B'$. Suppose that $\color{red}AB' = \frac{4}{3}$. The value of $\cos \theta$ can be written as $\frac{m}{n}$, where $m$ and $n$ are relatively prime positive integers. Find $m + n$.
\end{minipage}
\\
\midrule

Perturbed Variant &
\begin{minipage}[t]{12cm}\ttfamily
In a plane, there are points $A$ and $B$ such that $AB = 1$. Point $A$ is rotated around point $B$ counterclockwise through an acute angle $\theta$ to reach point $A'$. Next, point $B$ is rotated clockwise around point $A'$ through the same angle $\theta$, resulting in point $B'$. Given that $AB' = \frac{4}{3}$, express the value of $\cos \theta$ as $\frac{m}{n}$, where $m$ and $n$ are coprime positive integers. Determine $m + n$.
\end{minipage}
\\

\bottomrule
\end{tabular}

\caption{Examples of Target, Similar, and Perturbed Questions in Variable Renaming Perturbation.}
\label{tab:combined_questions_1}
\end{table*}
\FloatBarrier
\begin{table*}[t]
\centering
\small
\renewcommand{\arraystretch}{1.2}

\begin{tabular}{p{3cm} p{12cm}}
\toprule
\textbf{Settings} & \textbf{Content} \\
\midrule

Original Question &
\begin{minipage}[t]{12cm}
\ttfamily
A plane contains points $A$ and $B$ with $AB = 1$. Point $A$ is rotated in the plane counterclockwise through an acute angle $\theta$ around point $B$ to point $A^\prime$. Then $B$ is rotated in the plane clockwise through angle $\theta$ around point $A^\prime$ to point $B^\prime$. Suppose that $AB^\prime = \frac{4}{3}$. The value of $\cos \theta$ can be written as $\frac{m}{n}$, where $m$ and $n$ are relatively prime positive integers. Find $m+n$.

Present the answer in LaTeX format: $\boxed{\text{Your answer}}$
\end{minipage}
\\
\midrule

Similar Question &
\begin{minipage}[t]{12cm}
\ttfamily
A plane contains points $A$ and $B$ with $AB = \color{red}2$. Point $A$ is rotated in the plane counterclockwise through an acute angle $\theta$ around point $B$ to point $A^\prime$. Then $B$ is rotated in the plane clockwise through angle $\theta$ around point $A^\prime$ to point $B^\prime$. Suppose that $AB^\prime = \frac{5}{4}$. The value of $\cos \theta$ can be written as $\frac{m}{n}$, where $m$ and $n$ are relatively prime positive integers. Find $m+n$.
\end{minipage}
\\
\midrule

Perturbed Question &
\begin{minipage}[t]{12cm}
\ttfamily
A plane contains points $A$ and $B$ with $AB = \color{red}3$. Point $A$ is rotated in the plane counterclockwise through an acute angle $\theta$ around point $B$ to point $A'$. Then $B$ is rotated in the plane clockwise through angle $\theta$ around point $A'$ to point $B'$. Suppose that $AB' = \frac{7}{3}$. The value of $\cos \theta$ can be written as $\frac{m}{n}$, where $m$ and $n$ are relatively prime positive integers. Find $m + n$.
\end{minipage}
\\
\midrule

Perturbed Variant &
\begin{minipage}[t]{12cm}
\ttfamily
In a plane, there are points A and B such that the distance AB equals 3. Point A is rotated counterclockwise by an acute angle $\theta$ around point B to reach point $A'$. Next, point B is rotated clockwise by angle $\theta$ around point $A'$ to arrive at point $B'$. If the distance $AB'$ is $\frac{7}{3}$, then the value of $\cos \theta$ can be expressed as $\frac{m}{n}$, where $m$ and $n$ are relatively prime positive integers. Calculate $m + n$.
\end{minipage}
\\

\bottomrule
\end{tabular}

\caption{Examples of Target, Similar, and Perturbed Questions in Number Replacement Perturbation.}
\label{tab:combined_questions_2}
\end{table*}
\FloatBarrier
\begin{table*}[t]
\centering
\small
\renewcommand{\arraystretch}{1.2}

\begin{tabular}{p{3cm} p{12cm}}
\toprule
\textbf{Settings} & \textbf{Content} \\
\midrule

Original Question &
\begin{minipage}[t]{12cm}
\begin{ttfamily}
Isosceles triangle $\triangle ABC$ has $AB = BC.$ Let $I$ be the incenter of $\triangle ABC.$ The perimeters of $\triangle ABC$ and $\triangle AIC$ are in the ratio $125:6,$ and all the sides of both triangles have integer lengths. Find the minimum possible value of $AB.$

Present the answer in LaTeX format: $\boxed{\text{Your answer}}$
\end{ttfamily}
\end{minipage}
\\
\midrule

Similar Question &
\begin{ttfamily}
Isosceles triangle $\triangle DEF$ has $DE = EF.$ Let $\color{red}J$ be the incenter of $\triangle DEF.$ The perimeters of $\triangle DEF$ and $\triangle DEJ$ are in the ratio $100:5,$ and all the sides of both triangles have integer lengths. Find the minimum possible value of $DE.$
\end{ttfamily}
\\
\midrule

Perturbed Question &
\begin{ttfamily}
Isosceles triangle $\triangle DEF$ has $DE = EF.$ Let $J$ be the incenter of $\triangle DEF.$ The perimeters of $\triangle DEF$ and $\triangle DEJ$ are in the ratio $100:5,$ and all the sides of both triangles have integer lengths, {\color{red} as is common in many geometric problems.} Find the minimum possible value of $DE.$
\end{ttfamily}
\\
\midrule

Perturbed Variant &
\begin{ttfamily}
In isosceles triangle $\triangle DEF,$ the sides satisfy $DE = EF.$ Denote the incenter of $\triangle DEF$ as point $J.$ The ratio of the perimeters of $\triangle DEF$ to $\triangle DEJ$ is $100:5,$ and all sides of both triangles are integers, which is typical in geometric scenarios. Determine the smallest possible value for $DE.$
\end{ttfamily}
\\

\bottomrule
\end{tabular}

\caption{Examples of target, similar, and perturbed questions in distractor insertion perturbation.}
\label{tab:combined_questions_3}
\end{table*}
\FloatBarrier
\end{document}